\UseRawInputEncoding
\documentclass[journal]{IEEEtran} 

\IEEEoverridecommandlockouts
\usepackage{moreverb,url}

\usepackage[colorlinks,bookmarksopen,bookmarksnumbered,citecolor=red,urlcolor=red]{hyperref}

\usepackage{algorithm}
\usepackage{algpseudocode}
\usepackage{cite}
\usepackage{amsmath,amssymb,amsfonts}
\usepackage{graphicx}
\usepackage{textcomp}
\usepackage[table, dvipsnames]{xcolor}
\usepackage{float}
\usepackage{subcaption}
\usepackage{tabularray}
\usepackage{multirow}
\usepackage{booktabs}
\usepackage{hyperref}
\usepackage{titlesec}
\usepackage[disable]{todonotes}
\DeclareCaptionFont{mysize}{\fontsize{8}{10}\selectfont}
\usepackage{soul}
\usepackage[subtle,tracking=normal]{savetrees} 
\usepackage{balance}

\DeclareMathOperator*{\argmin}{arg\,min}

\newcommand\BibTeX{{\rmfamily B\kern-.05em \textsc{i\kern-.025em b}\kern-.08em
T\kern-.1667em\lower.7ex\hbox{E}\kern-.125emX}}

\def\BibTeX{{\rm B\kern-.05em{\sc i\kern-.025em b}\kern-.08em
    T\kern-.1667em\lower.7ex\hbox{E}\kern-.125emX}}

\newcommand{\multilinecomment}[1]{}

\author{Andrea Dal Prete*$^{1,2}$, Zeynep {\"O}zge Orhan*$^{1,3}, \IEEEmembership{Student Member, IEEE}$, \\Anastasia Bolotnikova$^{3,4}$ \IEEEmembership{IEEE Member, IEEE}, Marta~Gandolla$^{2}$, \\  Auke Ijspeert$^{3}$ \IEEEmembership{Fellow, IEEE}, and Mohamed Bouri$^{1,5}$ \IEEEmembership{Senior Member, IEEE}
\thanks{$^{1}$ REHAssist Group,  EPFL. $^{2}$ Department of Mechanical Engineering, Politecnico di Milano. $^{3}$ BioRobotics Laboratory (BioRob), EPFL. $^{4}$ Robotics and InteractionS Team, LAAS-CNRS. $^{5}$ Translational Neural Engineering Laboratory (TNE), EPFL.}
\thanks{(*) These two authors contributed equally to this study. Corresponding author e-mail: {andrea.dalprete@polimi.it}. This article has supplementary downloadable material available at
\url{https://zenodo.org/records/14076104}, provided by the authors. A video from the real-time experiments is available in \url{https://youtu.be/dxp4wHUmLrA}}
}

\begin{document}

%\runninghead{Dal Prete et al.}

\title{Locomotion Mode Transitions:\\ Tackling System- and User-Specific Variability in Lower-Limb Exoskeletons}

\maketitle

\begin{abstract}

Accurate detection of locomotion transitions, such as walk to sit, to stair ascent, and descent, is crucial to effectively control robotic assistive devices, such as lower-limb exoskeletons, as each locomotion mode requires specific assistance. Variability in collected sensor data introduced by user- or system-specific characteristics makes it challenging to maintain high transition detection accuracy while avoiding latency using non-adaptive classification models. In this study, we identified key factors influencing transition detection performance, including variations in user behavior, and mechanical design of the exoskeletons. To boost the transition detection accuracy, we introduced two methods for adapting a finite-state machine classifier to system- and user-specific variability: a Statistics-Based approach and Bayesian Optimization. Our experimental results demonstrate that both methods remarkably improve detection accuracy across diverse users, achieving up to an 80\% increase in certain scenarios compared to the non-personalized threshold method. These findings emphasize the importance of personalization in adaptive control systems, underscoring the potential for enhanced user experience and effectiveness in assistive devices. By incorporating subject- and system-specific data into the model, our approach offers a precise and reliable solution for detecting locomotion transitions, catering to individual user needs, and ultimately improving the performance of assistive devices.
\end{abstract}


\section{Introduction}

Effective assistance during locomotion activities such as walking, stair ascent, stair descent, and sitting is essential for enhancing user experience in lower-limb exoskeletons, because of this, accurate real-time transition detection is crucial for developing intelligent systems that provide responsive assistance. While most control strategies are designed for steady-state activities like walking, real-world exoskeleton applications need to adapt to diverse tasks and terrains, and variations in user gait and exoskeleton designs make reliable transition detection challenging, as existing methods often struggle in dynamic environments.

\begin{figure}[t]
    \centering
    \includegraphics[width=0.55\linewidth]{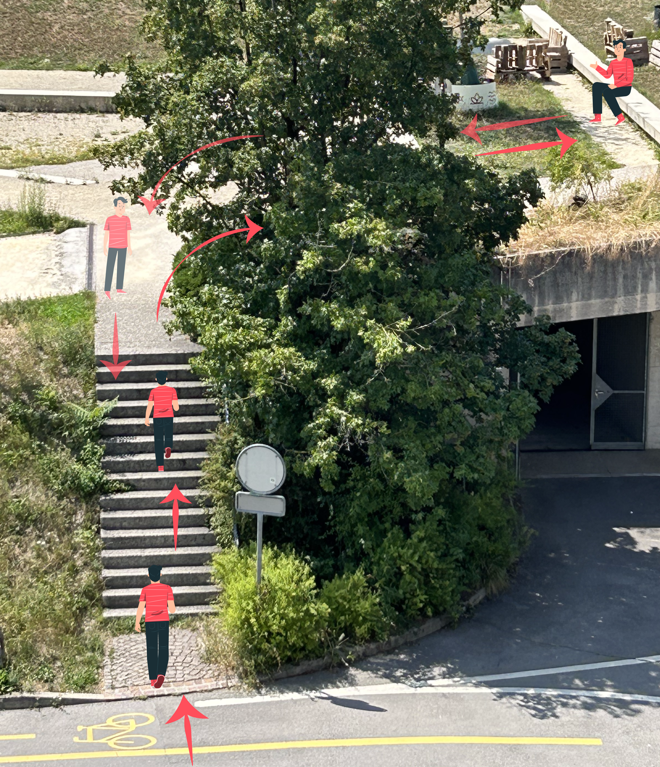}
\caption{The outdoor environment with transitions between locomotion modes.}
\label{fig:OnlineTestsOutdoorEnvironment}
\end{figure}

Locomotion mode identification involves recognizing various activities or movement patterns, such as walking, running, stair ascent, stair descent, and sitting \cite{Kolaghassi2021}. While existing methods excel in detecting steady-state activities \cite{Donahue2022, Gehlhar2023}, identifying transitions is critical for lower-limb exoskeletons to ensure seamless assistance, smooth human-robot interaction, enhanced safety and stability \cite{Cheng2021, Qian2022, Zhang2024, Haque2024, Wang2024, Cheng_2024}. This necessitates the development of specialized algorithms that focus on transition detection rather than continuous activity recognition. In this context, most studies on locomotion mode detection rely on body-mounted sensors but lack real-time integration with assistive devices \cite{Zhou2019, Lu2020_2, Narayan2022, Vu2022, Eken2023, Wang2024}. While prosthetic systems have achieved high real-time performance for recognizing locomotion mode transitions \cite{Young2014b, Su2019, Haque2024}, transition detection in lower-limb exoskeletons remains less explored \cite{Moreira2022}.

Long et al. \cite{Long2016} used an optimized support vector machine (SVM) for a knee exoskeleton to classify five modes, including transitions, achieving low error rates but limited real-time applicability. Wang et al. \cite{Wang2018} applied a Long Short-Term Memory network to a soft knee exoskeleton, attaining high accuracy but with recognition delays. Similarly, Zhou et al. \cite{Zhou2019} and Zhang et al. \cite{Zhang2024} explored machine learning (ML) approaches, but their reliance on fixed-length windows limits adaptability. Liu and Wang \cite{Liu2020} demonstrated real-time SVM-based recognition for a knee exoskeleton, while Du et al. \cite{Du2021} proposed a fuzzy logic-based approach for hip exoskeletons, avoiding subject-specific training. Kang et al. \cite{Kang2022} suggested a subject-independent deep learning model, though it remains untested in real-time scenarios. 

Most of the traditional ML methods effectively handle complex patterns but require extensive feature engineering, limiting scalability \cite{Alzubaidi2023}. Deep learning automates feature extraction and excels with large datasets but demands significant labeled data, high computational resources, and lacks interpretability, making it less suitable for assistive devices and human-machine interaction applications \cite{Alzubaidi2023}. To address these constraints, this study employs an ML-trained threshold-based (TH) method, enhancing generalizability without manual threshold setting. The framework suggested by Cheng et al. \cite{Cheng2021} was optimized in our earlier work to improve embedded system performance, reducing processing time by up to elevenfold \cite{Orhan2023_2}. The proposed ML-trained threshold-based method demonstrated high accuracy with ten subjects across walking, sitting, stair ascent, and stair descent in an outdoor setting as presented in Fig.~\ref{fig:OnlineTestsOutdoorEnvironment}. However, using this method with the autonomyo exoskeleton revealed certain limitations. While the method for detecting transitions between locomotion modes was generally effective across a diverse user base, we identified the need for subject and/or system-specific optimization to accommodate individual differences, especially when exoskeletons impose greater kinematic constraints on user movements. The more complex design of autonomyo, with three active degrees of freedom, affected user behavior differently than the lightweight eWalk, which has a single active degree of freedom. This complexity led to slower sitting patterns and pre-sit adjustments. Additionally, the ML-trained threshold-based method struggled with certain locomotion styles, like running-like stair climbing, highlighting the need for personalized transition detection. 

Inter-subject variability in response to assistive strategies suggests that device parameters beneficial to one individual may hinder another, highlighting the need for personalized control tuning \cite{Quesada2016, Kim2017_2, Slade2024}. Balancing personalization and generalization remains a core challenge in adaptive control systems for assistive devices. While initial personalization approaches relied on subject-specific training data, their practicality is limited by lengthy evaluation times \cite{Zhang2017, Ding2018, Xu2023}. To address this, Bayesian optimization has emerged as an efficient, noise-tolerant strategy, enabling human-in-the-loop (HIL) optimization to reduce user energy consumption and minimize experiment durations, critical for individuals with mobility impairments \cite{Brochu2010, Kim2017_2, Slade2024, Wen2020}. Building on prior research, our work leverages data from real-time experiments to improve transition detection accuracy by addressing system- and user-specific variability through a statistics-based approach and Bayesian optimization. Our findings show that personalized adjustments can remarkably improve the adaptability and effectiveness of lower-limb exoskeletons, paving the way for more responsive and user-centered assistive technologies. The major contributions of this study are the following:

\begin{itemize}
    \item This study evaluates an ML-trained threshold-based algorithm with two lower-limb exoskeletons, eWalk, and autonomyo, demonstrating its adaptability to varied design constraints.
    \item We introduce a statistics-based approach and Bayesian optimization to personalize transition thresholds, tailoring detection algorithms to individual gait patterns and system-specific characteristics. A method was also developed to address human-exoskeleton joint misalignment issues.
    \item We provide an open-source dataset of lower-limb kinematics, including data from 18 subjects (Ss) using two exoskeletons (eWalk and autonomyo) across four locomotion modes (walking, sitting, stair ascent, and descent) and their transitions.
\end{itemize}
\smallskip

\emph{Outline:} 
Section~\ref{sec:Methods} describes the feature space, finite-state machine, dataset construction, algorithms, and threshold-tuning methods. Section~\ref{sec:Experiments} details human subject experiments, with results in Section~\ref{sec:Results} and discussion in Section~\ref{sec:Discussion}. Section~\ref{sec:Conclusion} concludes the paper, while the Supplementary Material provides technical details on the shared dataset.

\section{Methods}
\label{sec:Methods}

\subsection{Training Dataset and Features}
\label{sec:Method-Dataset}

We use the dataset provided by Reznick et al. \cite{Reznick2021}, featuring activities such as walking, stair ascent/descent, sitting, and standing, with transitions noted as W-S, S-W, W-SA, SA-W, W-SD, and SD-W, for offline classifier training, by splitting it into 90\% for training/validation and 10\% for testing. Following \cite{Cheng2021}, instantaneous characteristic features (ICFs) are derived from thigh angle ($\theta_{th}$) and velocity ($\dot{\theta}_{th}$) to detect transitions within a gait step. Three ICFs (ICF-1, ICF-2, ICF-3) are used to classify transitions between Walk and Sit, Walk and Stair Ascent, and Walk and Stair Descent at three moments: (1) maximum hip flexion (MHF), (2) heel strike (HS), and (3) thigh angle within a specified range (THR), defined as $\theta{th} \in [62,75]_{\text{ewalk}}$ and $[55,70]_{\text{autonomyo}}$. ICF-1 is $\theta_{th}$ at MHF, ICF-2 is $\theta_{th,MHF} - \theta_{th,HS}$, and ICF-3 is $\dot{\theta}_{th}$ when $\theta_{th} \in THR$ as presented in Table~\ref{tab:ICFDefinitions}.

\begin{table}[t]
\centering
\caption{The table provides thresholds (Thr.) for each feature on the thigh angle trajectory ($\theta_{th}$) in $deg$ and thigh angular velocity ($\dot{\theta}_{th}$) in $deg/s$ for the eWalk and autonomyo (auto.) exoskeletons.}
\label{tab:ICFDefinitions}
\begin{tabular}{p{1.1cm} p{0.7cm} p{4.0cm} p{0.6cm} p{0.6cm}}
\toprule
Transition & Class. feat. & Definition & Thr. eWalk & Thr. auto. \\ 
\midrule
W-S      & ICF-3 & $\dot{\theta}_{th}$, when $\theta_{th} \in THR$ & 23.32 & 23.32 \\ 
S-W & ICF-3 & $\dot{\theta}_{th}$, when $\theta_{th} \in THR$ & -4.32 & -4.32 \\ 
W-SA & ICF-1 & $\theta_{th}$ at MHF $\theta_{th,MHF}$ & 50.52 & 50.52 \\ 
SA-W & ICF-1 & $\theta_{th}$ at MHF $\theta_{th,MHF}$ & 51.21 & 51.21 \\ 
W-SD & ICF-2 & The difference $\theta_{th,MHF}$ - $\theta_{th,HS}$ & 10.37 & 13.37 \\ 
SD-W      & ICF-2 & The difference $\theta_{th,MHF}$ - $\theta_{th,HS}$ & 9.62 & 9.62\\ 
\bottomrule
\end{tabular}
\end{table}

We use the FSM \cite{Cheng2021} with four states as Sit, Walk, Stair Ascent, and Stair Descent to address six transition detection scenarios, as shown in Fig.~\ref{fig:OnlineTestsOutdoorEnvironment}. The FSM combines level walking, standing, and ramp walking into a unified state, it relies on detectors to identify key gait moments (i.e. MHF and HS) based on thigh angle and ground reaction force data, and detects locomotion transitions by solving a binary classification problem on a single input ICF.

\subsection{Classification Algorithms and ML-Trained TH-Method}
Several machine learning algorithms (e.g., Support vector machines (SVM), Logistic Regression (LR), k-Nearest Neighbors, Na{\"i}ve Bayes (NB), Random Forest (RF), Gradient Boosting (GB)) were used to train classifiers offline using the Scikit-learn (v1.3.0) library in Python (v3.11.4) \cite{Scikit2011}. Hyperparameters were tuned heuristically, with the best-performing algorithms, Linear SVM and LR, selected for the FSM transition classifiers \cite{Orhan2023_2}. For NB, a Gaussian kernel was used, while KNN was configured with $k=5$ and Euclidean distance. RF and GB were set with $12$ and $2$ minimum splits, and $30$ and $100$ estimators, respectively, with GB using a learning rate of $\alpha=0.1$, while default settings were used for SVM and LR. To meet the computational constraints of embedded systems, we implemented an ML-trained TH-method to replace ML models with simple $if/else$ conditions derived from offline training \cite{Orhan2023_2}. This approach enhances real-time performance, reduces costs, and improves interpretability while maintaining accuracy. By using thresholds set through ML classifiers in a 1-D feature space \cite{Cheng2021}, we ensure efficient operation for assistive devices.

\subsection{Hardware}
\begin{figure}[b!]
    \centering
    \includegraphics[width=0.5\linewidth]{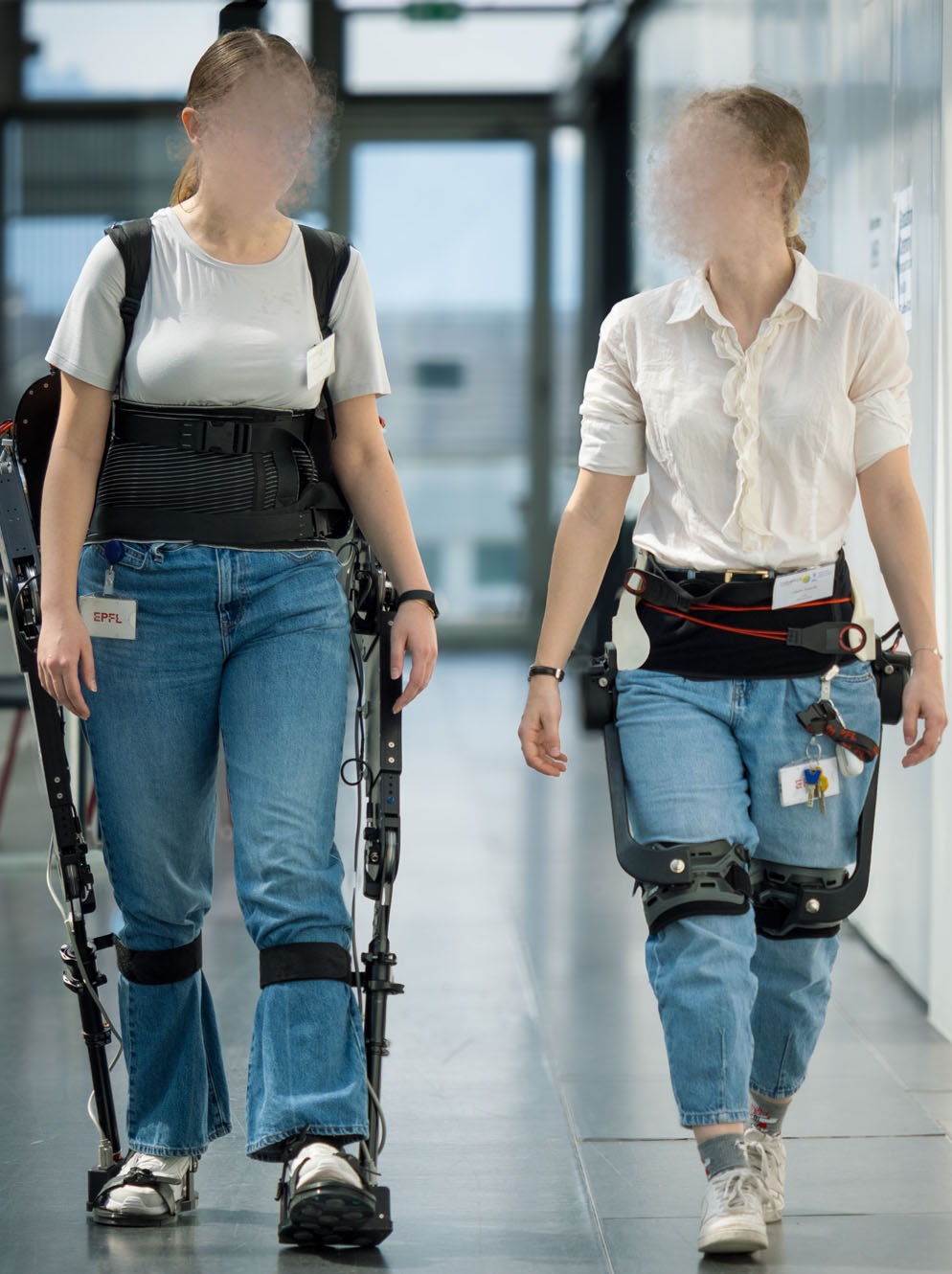}
\caption{The lower-limb exoskeletons autonomyo (left) and eWalk (right).}
\label{fig:exophoto}
\end{figure}

In this study, we used two lower-limb exoskeletons: eWalk and autonomyo, developed at EPFL REHAssist. The eWalk (eWalk v2), in collaboration with Sonceboz \cite{Messara2023}, assists with hip flexion/extension using 2 active degrees of freedom (DOF) actuated by DC servo motors while allowing natural walking with hip abduction/adduction and internal rotation Fig.~\ref{fig:exophoto}. The autonomyo exoskeleton (autonomyo v2), developed with Faulhaber Drive Systems \cite{Ortlieb2017, orhan2023}, is meant to assists individuals with walking impairments. It has 3 active and 3 passive DOF per leg, including hip (abd./add., flex./ext.) and knee (flex./ext.) joints, and passive ankle movements. It mounts brushless motors and a corresponding gearbox (3274 BP4 \& 42GPT, Faulhaber AG, Switzerland) with a 108:1 transmission ratio and additional cable and ball screw transmissions for flexion/extension and abduction/adduction. Torque sensors are integrated for hip and knee flexion/extension, and joint positions are measured using absolute encoders, with the zero position set by the homing technique. The torso angle is measured using an IMU (MPU6050) on autonomyo. Ground reaction forces (GRFs) are measured using force sensitive resistors (FSR) on eWalk and load cells on autonomyo. Both exoskeletons feature an embedded computer (BeagleBone Black) to log sensory data, with wireless communication via a Wi-Fi module.

\subsection{Human-Exoskeleton Joint Misalignment}

Human-exoskeleton joint angle discrepancies often arise from kinematic constraints and misalignments like rotational offsets and shifts in the center of rotation. While ergonomic and self-aligning mechanisms attempted to address these issues \cite{Schiele2006, Stienen2009, Celebi2013, Gálvez2016}, mechanical design improvements are not always feasible. Hence, we propose a regression-based mapping algorithm to correct misalignments and improve data accuracy, focusing on the hip flex./ext. joint in autonomyo:
\begin{equation}
    f(\pmb{w}, \pmb{\phi}(x)) = \pmb{\phi}(x)\pmb{w} \rightarrow t
\end{equation}
The algorithm aligns measured joint angles with literature data \cite{Reznick2021}, with $\pmb{w}$ the trainable weights, $\pmb{\phi}$ the basis functions, and $t$ the corrected angle. We formulate a sum of squares loss function between model output and literature data to optimize $\pmb{w}$:
\begin{equation}\begin{split}
L(\pmb{w})& = \left( \pmb{t} - f(\pmb{w}, \pmb{\phi}(\pmb{x}))\right)^T\left( \pmb{t} - f(\pmb{w}, \pmb{\phi}(\pmb{x}))\right)\\ 
=& \left( \pmb{t} - \pmb{\phi}(x) \: \pmb{w} \right)^T\left( \pmb{t} - \pmb{\phi}(x) \: \pmb{w} \right) 
\end{split}\end{equation}
And we minimize it with respect to the free parameters $\pmb{w}$:
\begin{equation}\label{eq:3}
    \nabla_{\pmb{w}} L(\pmb{w}) = 0 \longrightarrow \pmb{w}_{opt} = \left[\pmb{\phi}(\pmb{x})^T\pmb{\phi}(\pmb{x})\right]^{-1}\pmb{\phi}(\pmb{x})^T\pmb{t}
\end{equation}

The final function accounts for position and its first and second derivatives as inputs:
\begin{equation}
    f(x, \dot{x}, \ddot{x}) \rightarrow t, \quad f: \mathbb{R}^3 \rightarrow \mathbb{R}
\end{equation}
where $\pmb{w} = [w_0, w_1, w_2, w_3, w_4, w_5, w_6]^T$ are the trainable parameters coupled with the linear and nonlinear combinations of position, velocity, and acceleration. Specifically:
\begin{equation}
    \pmb{\phi}(x) = [1, x, \dot{x}, \ddot{x}, x \cdot \dot{x}, x \cdot \ddot{x}, \dot{x} \cdot \ddot{x}]
\end{equation}
At each time step, we combine the encoder's hip position ($x$), velocity ($\dot{x}$), and acceleration ($\ddot{x}$) to produce a corrected hip angle ($t$). The model, with 7 trainable parameters ($\pmb{w}$), is trained on 12 W-SA transition cycles from 10 subjects, using literature means as ground truth (GT). We evaluate training performance with root mean squared error (RMSE), generalizability across subjects, and visual mapping confirmation. We speculate that the nonlinear flexibility of $\pmb{\phi}$ enables the model to handle data nonlinearity, providing a system-agnostic solution.

\subsection{User/System Specific Optimization Methods}
\label{sec:SSspecific}

The proposed ML-trained TH-method balances computational efficiency and generalization for transition detection but accuracy may decrease when handling gait data from subjects significantly different from the training population. To mitigate this, the following sections present fine-tuning techniques to improve detection accuracy, particularly for outlier subjects.

\subsubsection{Statistics-Based Tuning}

\begin{figure}[b]
    \centering
    \includegraphics[width=0.8\linewidth]{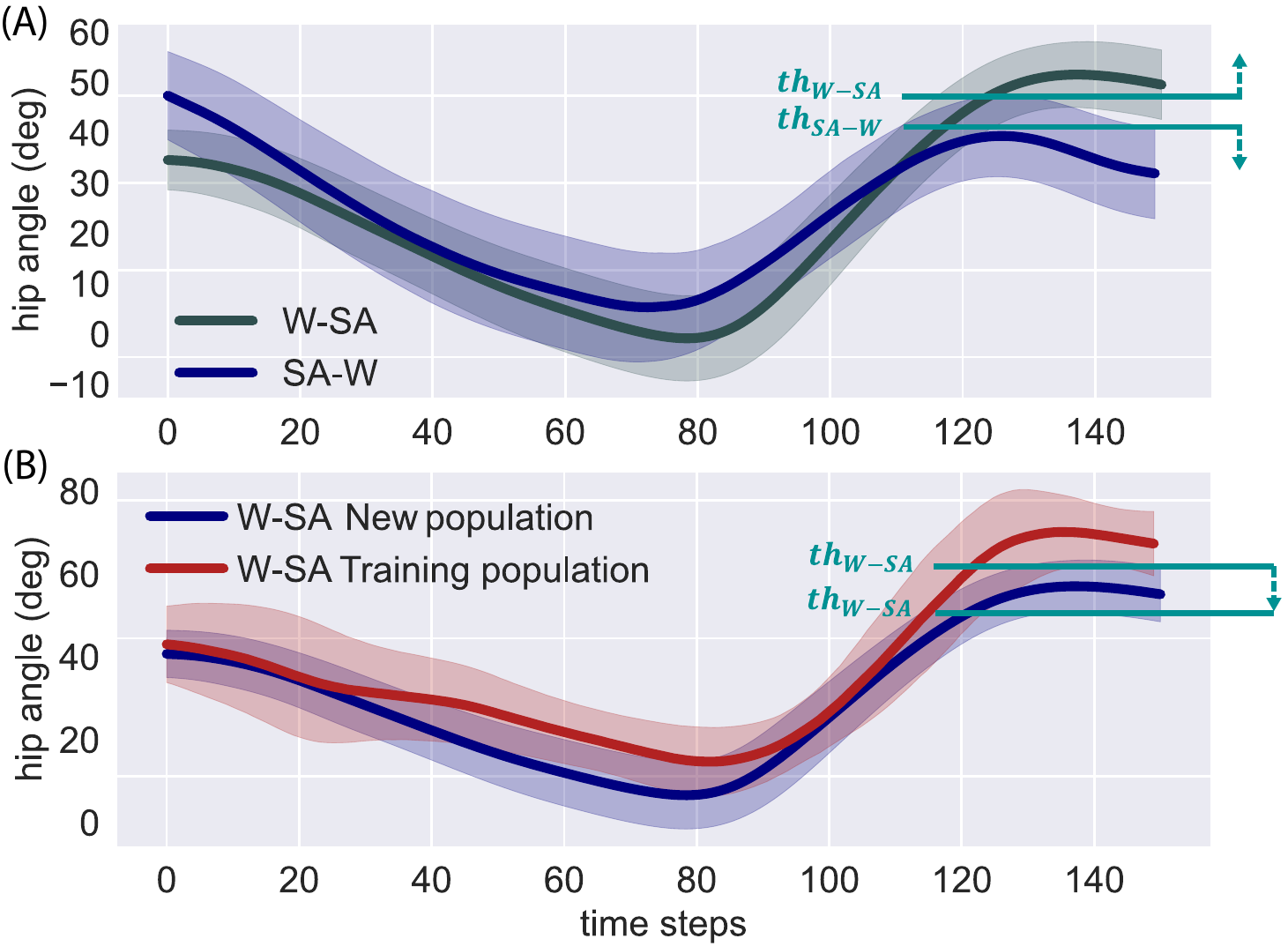}
    \caption{Visualisation of the upper and lower bounded transitions.}
    \label{fig:UpperLowerBounded}
\end{figure}  

We developed the \textit{Statistics-Based approach} (SBA) to adapt thresholds by analyzing statistical differences between training and subject-specific data. The FSM uses these thresholds as upper and lower bounds to detect transitions, such as in SA-W and W-SA, where the ICF must exceed or fall below trained thresholds ($th_{W-SA}$ and $th_{SA-W}$) for detection, as shown in Fig.~\ref{fig:UpperLowerBounded}-A, and this approach applies to all transition pairs. Fig.~\ref{fig:UpperLowerBounded}-B shows a scenario where the trained threshold is too high for the new gait pattern, causing missed detections. To improve performance, we propose lowering the thresholds based on the new population's statistical properties, adjusting the training thresholds as defined by the following formula:

%\vspace{-0.3cm}
\begin{equation}\label{eq:statistics_based}
    TH_{new} = TH_{tr} \cdot \frac{mean(ICF_i^{new})\pm std(ICF_i^{new})}{mean(ICF_i^{tr})\pm std(ICF_i^{tr})}
\end{equation}

Where $TH_{tr}$ is the old threshold from training, and $TH_{new}$ is the adjusted threshold after tuning. A $+$ sign is used for lower bound transitions, and a $-$ sign otherwise. $ICF_i^{new/tr}$ denotes the feature for the specific transition cycle in the new or training data. Summing up, we extract $ICF_i$ for both populations and use Eq.~\ref{eq:statistics_based} to scale the threshold. 

\subsubsection{Bayesion Optimization-based Tuning} 

Although SBA allows us to scale transition thresholds so subject-specific statistics, formally the thresholds can be adjusted by solving an optimization problem:

%\vspace{-0.3cm}
\begin{equation}\label{eq:ThresholdsOptimization}
%\vspace{-0.2cm}
\begin{aligned}
\pmb{TH}^* = & \:\:\: \argmin_{\pmb{TH} \in \mathbb{R}^n} \:\:\: J(\pmb{TH}) \\
& \text{subject to} \:\:\:\:\: \pmb{TH} \in \Omega
\end{aligned}
\end{equation}

Here, $J$ is some objective function measuring FSM performance, and $\pmb{TH}$ denotes the optimization thresholds. We use \textit{Bayesian Optimization} (BO) to fine-tune these thresholds when the impact of threshold changes on performance is unclear and the explicit formulation of $J(\pmb{TH})$ is unavailable \cite{Garnett_2023}. BO iteratively samples the parameter-output relationship and uses an \textit{acquisition function} to guide optimization with minimal experiments. For objective function estimation, we use Gaussian Processes with a Radial Basis Function kernel via the GPy library in Python \cite{GPy}, chosen for its smoothness, higher-order derivatives, as well as the capability to provide both mean and uncertainty values from data fitting \cite{Rasmussen2006Gaussian}. From the GP fitting we derive a \textit{surrogate function} $S(\pmb{TH})$, which we use to formulate the acquisition function as follows:
\begin{equation}
    A(\pmb{TH}) = k*mean(S(\pmb{TH})) \pm (1-k)*std(S(\pmb{TH}))
    %\vspace{-0.15cm}
\end{equation}

The parameter $k$ in BO balances exploration and exploitation. Since our search space has multiple local minima, we set $k=0.5$ to prioritize balanced exploration and rapid convergence. The acquisition function $A$represents an estimate of the objective function and reflects the relationship between parameters and performance. BO iteratively refines the surrogate function with new and past data and minimizes the acquisition function to find optimal thresholds, recursively solving the following instead of Eq.~\ref{eq:ThresholdsOptimization}:
\begin{equation}
\begin{aligned}
\pmb{TH}^* = \:\:\: & \argmin_{\pmb{TH} \in \mathbb{R}^n} \:\:\: A(\pmb{TH}) \\
& \text{subject to} \:\:\:\:\: \pmb{TH} \in \Omega
\end{aligned}
\end{equation}

Algorithm \ref{alg:pseudocode1} provides a pseudocode of the BO process. We optimize thresholds in pairs, and the searching space $\Omega$ per each one is set $TH_{WtS/StW}\:\: \in \:\: [-60,60]$, $TH_{WtSA/SAtW} \:\: \in \:\: [30,65]$, $TH_{WtSD/SDtW}\:\: \in \:\: [0,25]$).

\begin{algorithm}
\caption{Bayesian Optimization Process}\label{alg:pseudocode1}
\begin{algorithmic}[1]
\State \textbf{Variables:} $J$, $\pmb{TH}$
\While{not convergence condition}
    \State $J,\pmb{TH} \gets new\_sample$ \Comment{from experiment/simulation}
    \State estimate $J(\pmb{TH})$ \Comment{By fitting $\{J,\pmb{TH}\}$ pairs with GP}
    \State formulate $A(\pmb{TH})$
    \State suggest $\pmb{TH}_{\text{next}}$ from $\pmb{TH}_{\text{next}} = \arg \min_{\pmb{TH}} A(\pmb{TH})$
\EndWhile
\end{algorithmic}
\end{algorithm}

We formulate the objective function $J(\pmb{TH})$ as detailed in the pseudocode given in Algorithm~\ref{alg:BO_objectivefuntion}.

\begin{algorithm}
\caption{Objective function formulation} \label{alg:BO_objectivefuntion}
\begin{algorithmic}[1]
\If{$\text{GT} = class1$ \textbf{and} $\text{FSM\_output} \neq class1$}
    \State $J_{j+1}(\pmb{TH}) \gets J_j(\pmb{TH}) + C_1$
\EndIf

\If{$transition\_to\_class1\_happened\_before$}
    \If{$\text{GT} = Walk$ \textbf{and} $\text{Output\_FSM} \neq Walk$}
    \State $J_{j+1}(\pmb{TH}) \gets J_j(\pmb{TH}) + C_2$
\EndIf
\EndIf
\end{algorithmic}
\end{algorithm}

Where $class1$ refers to the classes the algorithm can detect from ``Walking'' (e.g., S, SA, SD). $GT$ suggests the requirement of supervised experiments with labeled data for BO. The penalty constants $C_1$ and $C_2$ are heuristically set to $0.005$ and $0.001$ respectively, with $C_1$ being higher to reduce the risk of getting stuck in a local minimum where the system prioritizes the recognition of the second transition. By assigning a higher penalty to the first transition in a sequence, we ensure it is recognized first. Thus, the FSM is penalized whenever it outputs a class different from the ground truth, with greater penalties the longer the time the FSM takes to detect the transition. Furthermore, to ensure optimized thresholds remain within realistic bounds, we add additional regularization by adding a penalty term to the objective function $J_j$ when thresholds assume values too far away in the searching space. For W-SA/SA-W transitions, a penalty is applied if any threshold exceeds $55^{\circ}$, while for W-SD/SD-W transitions, the penalty is applied when thresholds fall below $5^{\circ}$. In both cases the penalty is proportional to the squared Euclidean distance from the \textit{limit values}. The penalty is scaled by a regularization factor $\alpha$: \begin{equation} J_j = J_j + \frac{\alpha}{2}||\pmb{TH} - [\text{limit}, \text{limit}]||^2_2 \end{equation} where the \textit{limit values} are $55^{\circ}$ and $5^{\circ}$ for the respective transitions, $\pmb{TH} = \left[TH_{\text{transit1}}, TH_{\text{transit2}}\right]$, and $\alpha$ is a parameter governing the strength of the regularization, with no regularization when $\alpha = 0$. In our case, we heuristically set $\alpha=2.10^{-5}$. This regularization term, inspired by neural network weight decay strategies \cite{Chollet2021}, helps prevent getting stuck in local minima while expanding the search space for optimal solutions. As detailed in Section \ref{sec:Experiments}, our experimental scenario included 5 transitions per subject. We applied BO to subjects and transitions with poor performance, using 2 transitions for training and 3 for evaluation, and optimizing over 30 iterations. To validate the effective exploration and accurate objective estimation, we conducted grid search in parallel to BO.

\subsubsection{Discrete Grid Search Based Tuning}

The discrete Grid Search algorithm (GSA) is used as a baseline method to optimize thresholds due to its simplicity and ease of implementation. GSA evaluates a fixed number of conditions and selects the one with the lowest cost. The design space $\Omega$ is the same of BO: $TH_{W-S/S-W} \in [-60,60]$, $TH_{W-SA/SA-W} \in [30,65]$, $TH_{W-SD/SD-W} \in [0,25]$. For W-SA/SA-W and W-SD/SD-W, the space is discretized in 2.5-degree increments, and for W-S/S-W in 10-degree increments. The same cost function is evaluated at each point, providing a baseline for comparison with BO.

\section{Real-Time Experiments with eWalk and autonomyo}
\label{sec:Experiments}

The experiment focused on the four locomotion modes level walking, stair ascent, stair descent, and sitting and the six transitions between them (W-SA, SA-W, W-S, S-W, W-SD, SD-W) in the environment shown in Fig~\ref{fig:OnlineTestsOutdoorEnvironment}. Subjects completed 15-step stairs and selected their walking speed and sitting position, with each transition repeated 50 times per exoskeleton, given the five trials per subject. Eligible participants were healthy adults aged 18-65, with a height of 160-200 cm and EU shoe size 38-45. Exclusion criteria included systemic disorders, psychiatric impairments, and pregnancy. The eWalk experiment included 10 subjects (age $25.6 \pm 2.76$ years, height $178.4 \pm 8.1$ cm, weight $71.9 \pm 9.68$ kg), and the autonomyo test included 10 subjects (age $26.9 \pm 3.42$ years, height $178.2 \pm 8.65$ cm, weight $72.2 \pm 13.44$ kg). All participants provided written consent, in line with ethical standards set by the EPFL Human Research Ethics Committee and the Helsinki Declaration. During the tests, eWalk and autonomyo operated in zero-torque transparent mode. Ground truth data were labeled in real-time by an observer, noting the transition and steady-state locomotion type.

\section{Results}
\label{sec:Results}

\subsection{Joint Trajectories \& Effect of Joint Misalignment}

\begin{figure}[t!]
    \centering
    \includegraphics[width=0.95\linewidth]{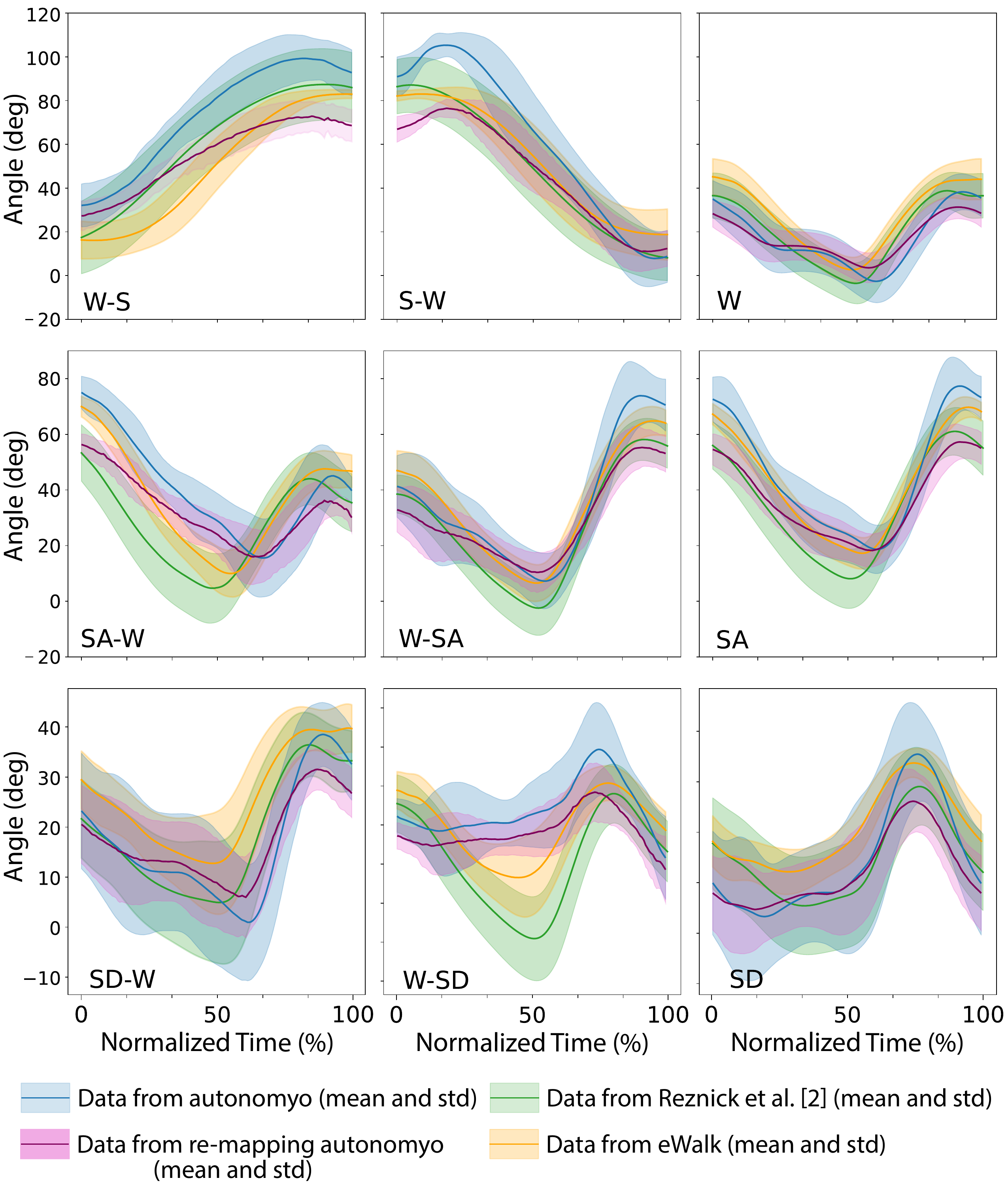}
    \caption{Hip/thigh angle trajectories from measurements of eWalk and autonomyo experiments, respectively, for tested locomotion modes with the re-mapped autonomyo joint angles compared to literature data given in \cite{Reznick2021}.}
    \label{fig:thighAngleTrajectories}
\end{figure}

In this study, we aimed to capture natural human movement using joint angle measurements. Fig.~\ref{fig:thighAngleTrajectories} presents the hip and thigh angle trajectories from experiments with the eWalk and autonomyo exoskeletons. The labeled time-series data from the experiments are available in \cite{dataset2024}, along with an example script for dataset exploration. We compare the trajectories with literature data from \cite{Reznick2021}, and notice strong alignment with established biomechanical patterns. Nevertheless, the data from autonomyo experiments displayed joint misalignment issues between the exoskeleton and the human hip joint in some scenarios. To address this, we train the mapping function on the dataset $D = \{\pmb{x},\pmb{t}\}$, where $\pmb{t}$ consists of W-SA transition cycles from the literature and $\pmb{x}$ comes from autonomyo. The resulting optimal weights are presented in Table~\ref{tab:MappingWeights}. We avoid normalization to minimize unnecessary computations, resulting in the different scales of the weights.

\begin{figure}[t]
    \centering
    \includegraphics[width=0.9\linewidth]{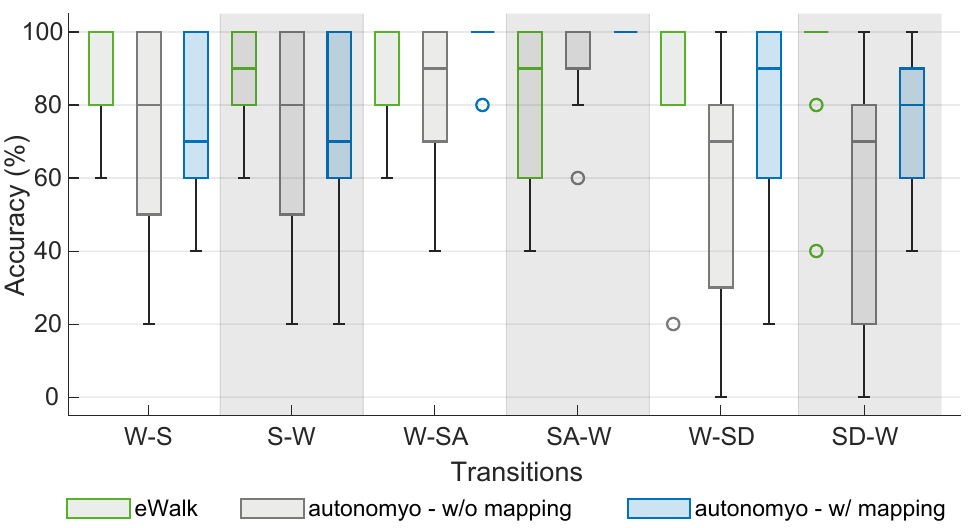}
    %\vspace{-0.2cm}
    \caption{Real-time recognition accuracy (\%) of locomotion transitions from experiments with eWalk and autonomyo exoskeletons.}
    %\vspace{-0.2cm}
\label{fig:RecognitionAccuracyTransitions2Exo_wandwoMapping}
\end{figure}

\begin{table}[h]
\centering
\caption{Mapping function's weights outcome from the training}
\label{tab:MappingWeights}
\begin{tabular}{lllllll}
\toprule
$w_0$ & $w_1$ & $w_2$ & $w_3$ & $w_4$ & $w_5$ & $w_6$ \\ 
\midrule
5.7 & 6.7e-1 & 3.3e-2 & 3.1e-4 & -4.6e-4 & 5.7e-6 & -2.5e-6 \\
\bottomrule
\end{tabular}
%\vspace{-0.05cm}
\end{table}

The mapping function reduced the mean MHF value from $73.87^\circ$ to $55.27^\circ$, aligning more closely with the literature value of $60.55^\circ$ \cite{Reznick2021}, and the mean standard deviation across gait cycles decreased from $17.48^\circ$ to $11.95^\circ$, compared to the literature value of $10.49^\circ$. At the same time, the mean RMSE between the literature data and the mapped joint angle was reduced by $3.61^\circ$. To validate the mapping function's generalizability, we applied it to other locomotion modes, as shown in Fig.~\ref{fig:thighAngleTrajectories}, registering reduced data variance and improved consistency in capturing natural human movement.

\subsection{Transition Detection Before \& After Optimization}\label{sec:ThresholdOptimization}
The recognition accuracy is defined as shown in Eq.~\ref{eq:transitionAccuracy}:

\begin{equation}
%\vspace{-0.1cm}
\label{eq:transitionAccuracy}
A = \frac{ N_{\text{CDT}}}{ N_{\text{TT}}} \cdot 100
\end{equation}
where $N_\text{CDT}$ is the number of transitions detected at correct moment and $N_\text{TT}$ is the total number of transitions. A transition is considered undetected if it is not recognized within a one-step delay, even if the steady state is correctly classified after a few steps. The accuracy is evaluated as binary, based on the detection condition. We excluded Ss 3 and 7 from autonomyo data due to discomfort during the experiment. The recognition accuracy results are shown in Fig.~\ref{fig:RecognitionAccuracyTransitions2Exo_wandwoMapping}. 

For the eWalk experiments, the median detection accuracies were: 100\%, 90\%, 100\%, 90\%, 100\%, and 100\% for W-S, S-W, W-SA, SA-W, W-SD, and SD-W, respectively. For autonomyo, the accuracies were 80\%, 80\%, 90\%, 100\%, 70\%, and 70\%. After applying the re-mapped joint angles, accuracies changed as follows: W-S and S-W decreased from 80\% to 70\%, W-SA increased from 90\% to 100\%, SA-W stayed at 100\%, W-SD increased from 70\% to 90\%, and SD-W improved from 70\% to 80\%. The standard deviation was reduced for all transitions except S-W.

\begin{figure}[b!]
    \centering
    \includegraphics[width=0.9\linewidth]{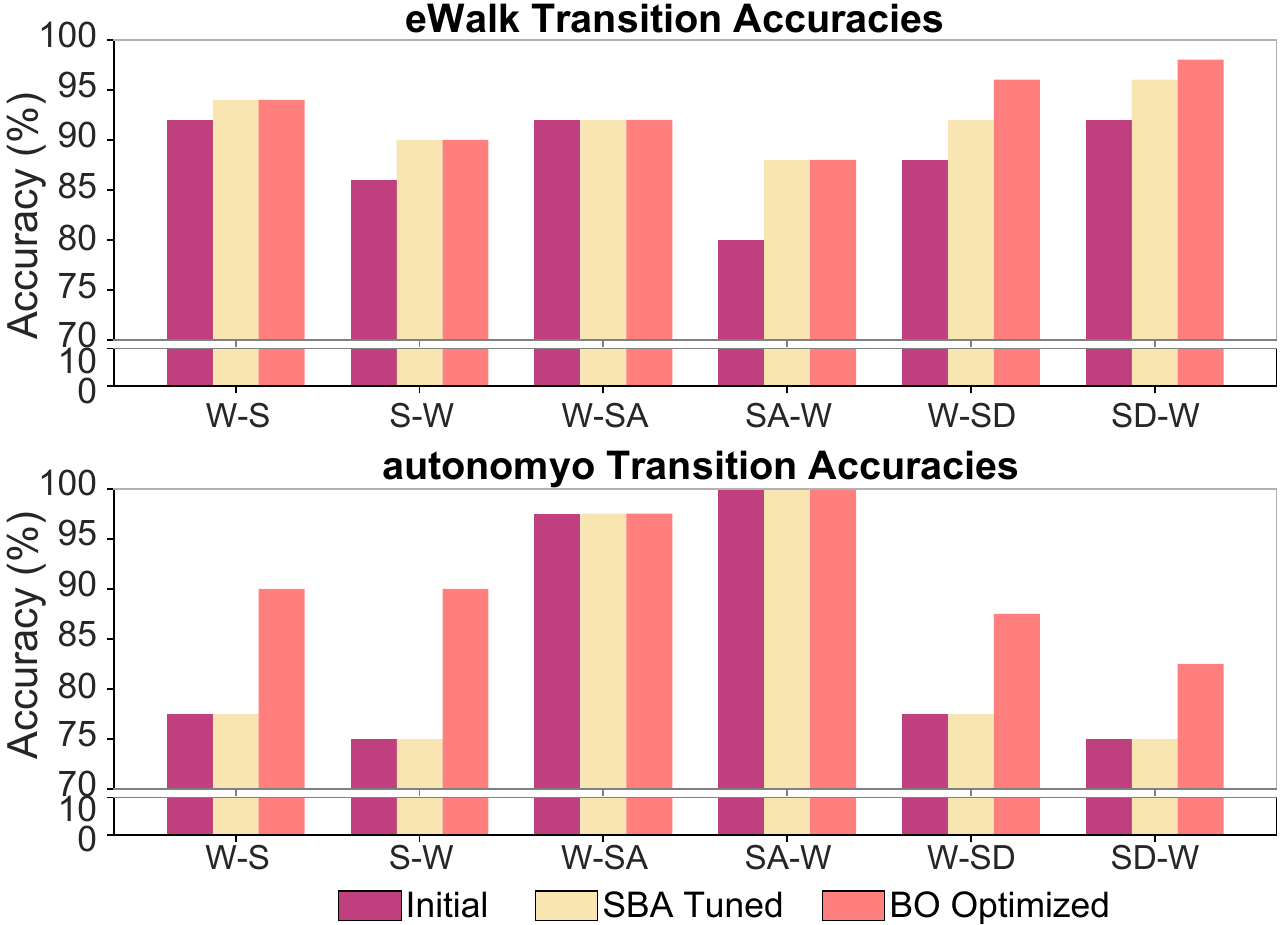}
    \caption{Changes of the transition detection accuracy results with SBA and BO for eWalk and autonomyo.}
    \label{fig:MeanAccuracyChanges_ewalk_autonomyo_initialvsSBAvsBO}
\end{figure}

Fig.~\ref{fig:MeanAccuracyChanges_ewalk_autonomyo_initialvsSBAvsBO} shows transition detection accuracies from eWalk and autonomyo experiments after using SBA and BO. For eWalk, both methods improved accuracy, with BO yielding slightly better results. Specifically, for eWalk, SBA improved transition accuracies for W-S from 92\% to 94\%, S-W from 86\% to 90\%, and SA-W from 80\% to 88\%. BO achieved the same increases for W-S, S-W, and SA-W, while further increasing W-SD accuracy to 96\% and SD-W accuracy to 98\%. For autonomyo, SBA showed no improvement, while BO enhanced accuracy across all transitions: W-S from 77.5\% to 90\%, S-W from 75\% to 90\%, W-SD from 77.5\% to 87.5\%, and SD-W from 75\% to 82.5\%.

\begin{figure*}[t!]
    \centering
    \includegraphics[width=1.0\linewidth]{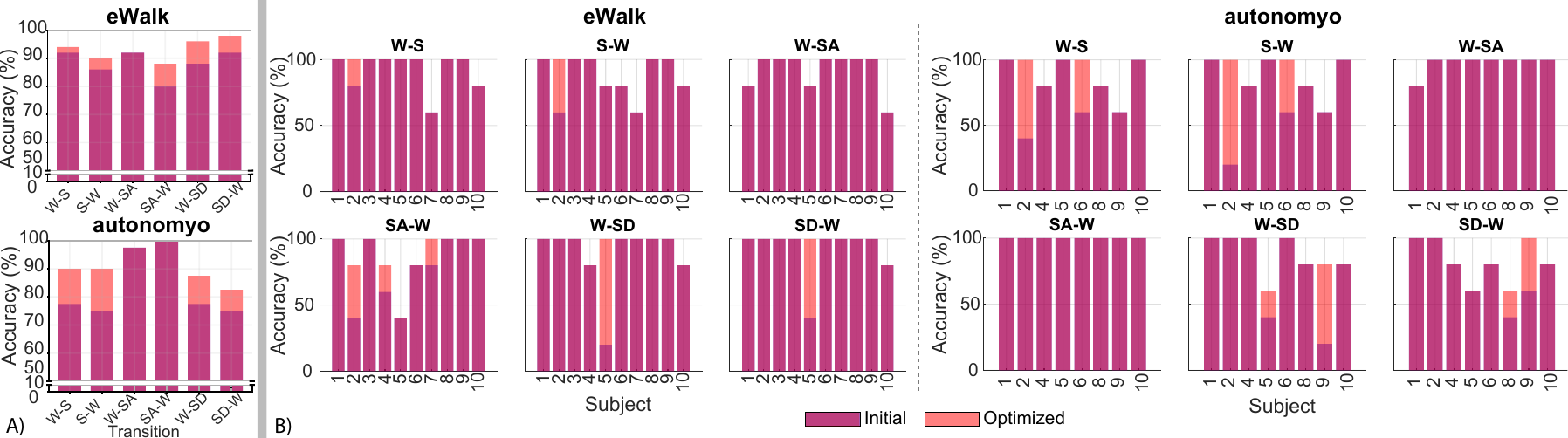}
    %\vspace{-0.7cm}
    \caption{A) Changes of the overall mean accuracy results with the subject-specific thresholds after BO for each of the transitions B) Changes of the accuracy results of each subject with the subject-specific thresholds after BO for each of the transitions for the eWalk and autonomyo experiments.}
    %\vspace{-0.1cm}
    \label{fig:SubjectSpecificChanges_ewalkandAutonomyo_v2}
\end{figure*}

The final accuracy results for both systems are presented in Fig.~\ref{fig:SubjectSpecificChanges_ewalkandAutonomyo_v2}, illustrating the effectiveness of BO at both group and individual levels. In eWalk experiments, BO notably improved accuracy for most subjects, adapting to individual gait patterns. For example, S2's W-S transition was missed before BO, but all transitions were correctly detected after optimization. S5 showed a significant increase in W-SD from 20\% to 100\% and SD-W from 40\% to 100\%, though no improvement was observed for the SA-W transition. The mean accuracy changes for eWalk demonstrate that BO improved overall detection rates and handled variability over subjects, suggesting more consistent performance across the user base.

For the autonomyo experiments, the results showed varied outcomes across subjects. While some subjects who initially had lower detection accuracies achieved substantial improvements following BO, others exhibited more moderate changes. For example, the S-W and W-S transition detection accuracy for S6 improved by 40\% after BO, reaching 100\% accuracy. However, more modest increases were observed for S5 during W-SD and S8 during SD-W, with only a 20\% improvement following threshold personalization. This variability suggests that BO enhanced transition detection accuracies across a diverse range of users, despite the differences in improvement levels, effectively adapting the system to each subject’s unique interaction with the exoskeleton, and optimizing transition detection based on individual characteristics. 

\begin{figure}[b]
    \includegraphics[width=0.98\linewidth]{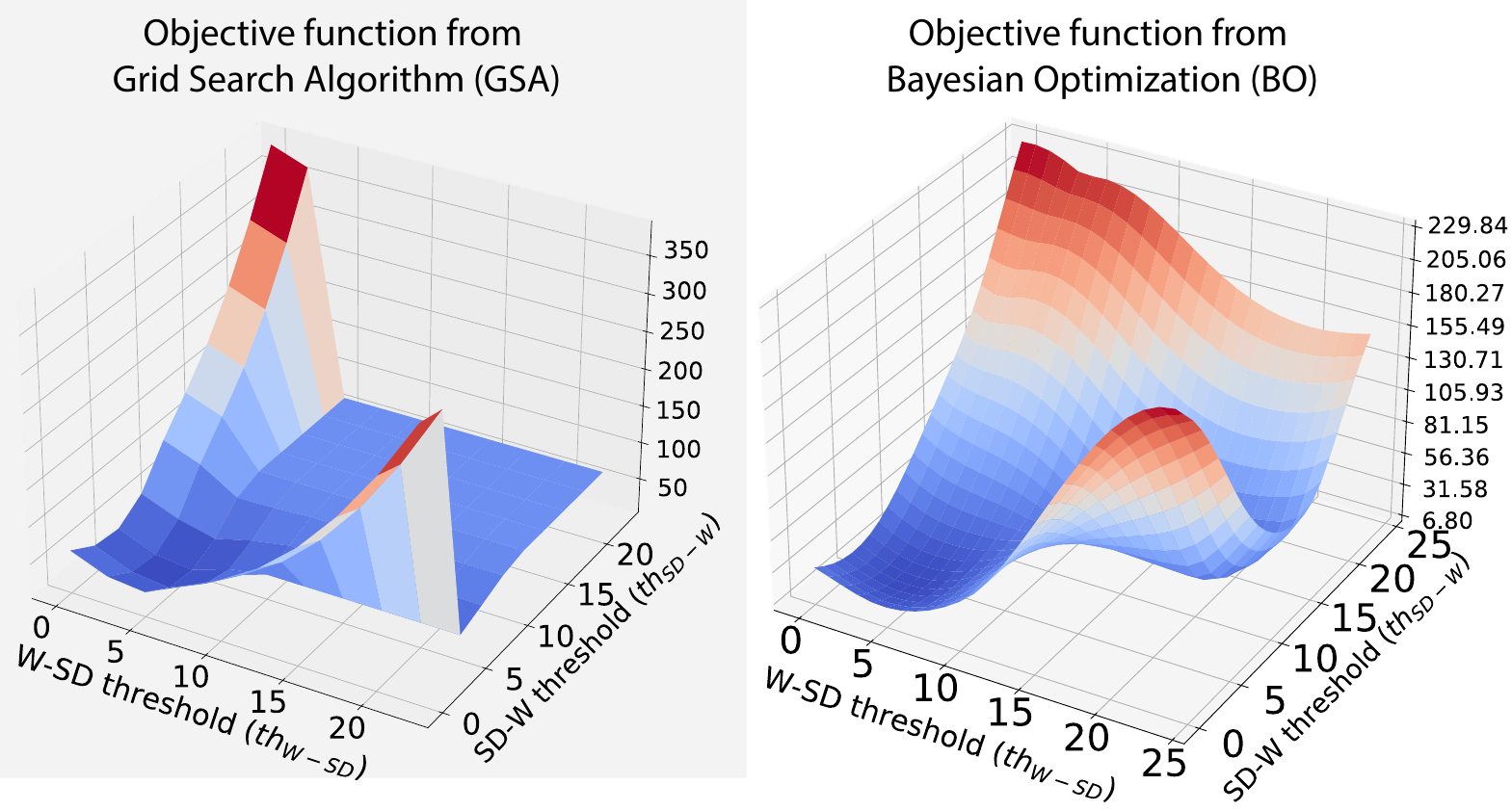}
    \caption{Acqusition function obtained by Grid Search Algorithm and Bayesian Optimization with sampling over iterations.}
    \label{fig:BayesianOptimizationProcess}
\end{figure}
%\vspace{-0.2cm}
\subsection{Comparison of Bayesian Optimization and Grid Search}

\begin{figure}[t]
    \centering
    \includegraphics[width=0.94\linewidth]{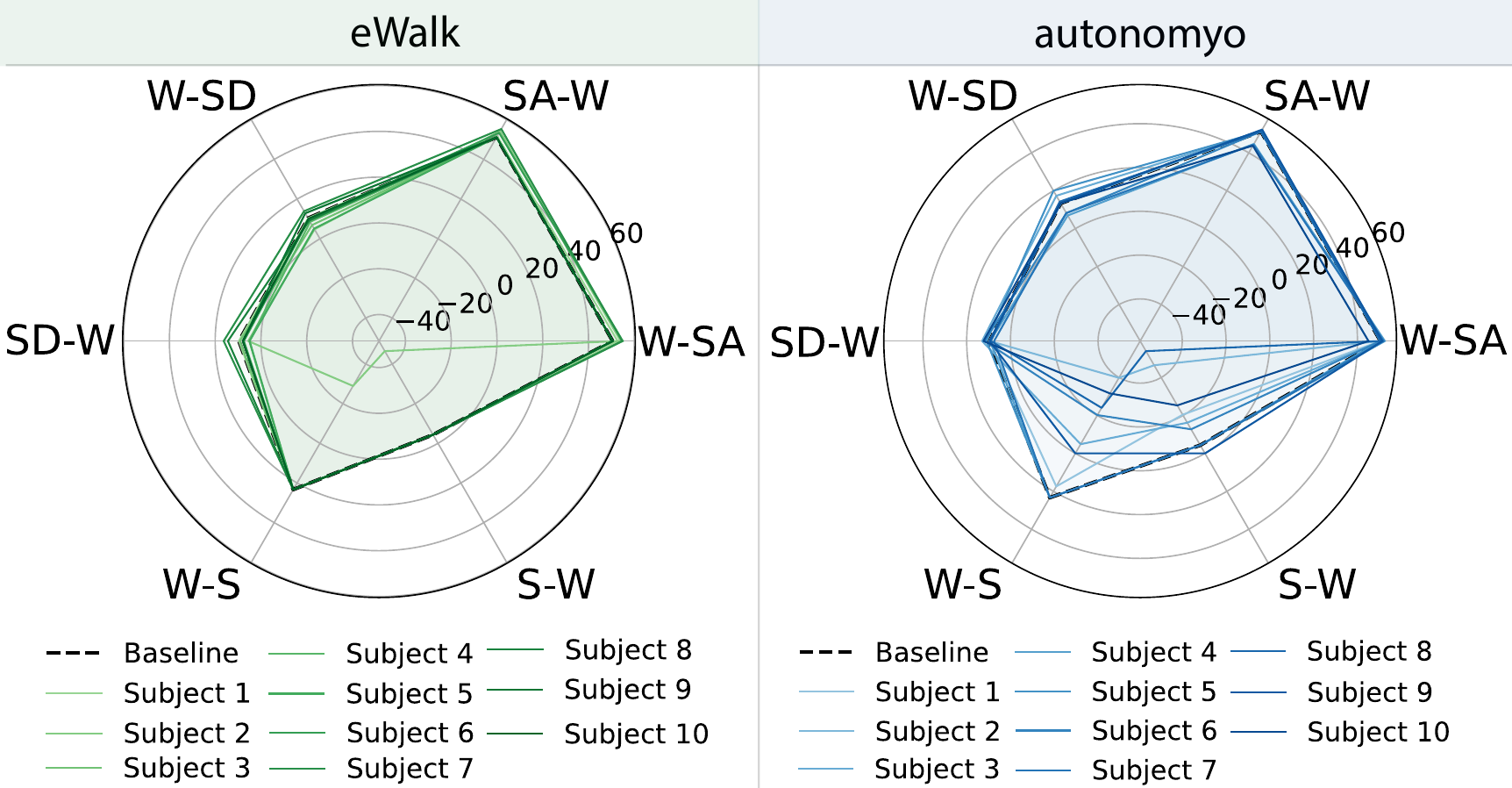}
    \caption{Changes in the thresholds after BO for each of the transitions for the eWalk and autonomyo experiments.}
    \label{fig:ThresholdChanges_Baseline_BO}
\end{figure}

Fig.~\ref{fig:BayesianOptimizationProcess} compares the BO process and GSA for optimizing the W-SD/SD-W transition thresholds for S5 using eWalk. The BO acquisition function closely mirrors the objective function of GSA, particularly at the global minimum. Both methods converge to similar thresholds, but BO achieves this in 30 iterations versus the 100 needed for GSA to evaluate the objective function. The final thresholds for each subject after BO optimization are summarized in Fig.~\ref{fig:ThresholdChanges_Baseline_BO}.

\begin{figure}[b!]
    \centering
    \includegraphics[width=0.82\linewidth]{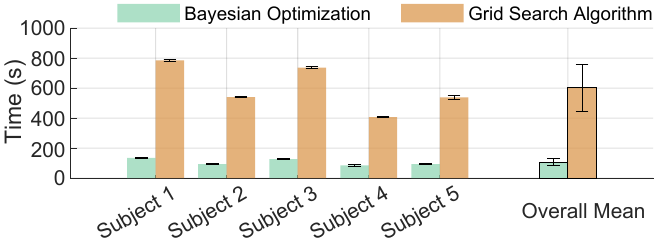}
    \caption{The computational time for the optimization with BO and GSA for the five subjects.}
    \label{fig:CompTime_BO_GSA_ewalk}
\end{figure}

In terms of computation cost, both BO and GSA were run 10 times per subject on the same machine with an Apple M1 processor (8-core CPU, 7-core GPU, and 8GB RAM) keeping track of inference time. Fig.~\ref{fig:CompTime_BO_GSA_ewalk} shows that, on average, GSA took four times longer than BO to optimize the thresholds, emphasizing BO's computational efficiency.

\section{Discussion}
\label{sec:Discussion}

\subsection{Public Dataset of the Joint Trajectories}

The joint trajectories from the eWalk and autonomyo exoskeletons demonstrate their potential to facilitate natural human movement across various locomotion modes. The alignment with literature \cite{Reznick2021} confirms the reliability of the exoskeleton measurements, particularly in the eWalk, where hip flex./ext. angles closely resemble natural human movement. This accuracy is essential for smooth human-robot interaction, enhancing user comfort and system effectiveness. Despite growing research, publicly available datasets for exoskeletons remain scarce, with most focusing on non-exoskeleton-assisted gait \cite{Moore2015,Fukuchi2018,Hu2018,Camargo2021,Reznick2021}. Our dataset\footnote{\url{https://zenodo.org/records/14076104}} addresses this gap by providing joint trajectory data from eWalk and autonomyo, serving as a valuable resource for evaluating and comparing assistive robotics control strategies.

\subsection{The Importance of Joint Alignment}

Accurate joint trajectory data is crucial for locomotion mode detection, but joint misalignment between human and exoskeleton joints remains a challenge \cite{Naf2019, Siviy2023}. Misalignment, often caused by simplified kinematic models or improper fitting \cite{BesslerEtten2022}, reduces control accuracy and user comfort. While eliminating kinematic mismatches is often tough \cite{Naf2019}, our approach minimizes discrepancies between exoskeleton measurements and human movement, enhancing system performance. The learning-based mapping function improves the alignment of joint angles with natural transitions, such as MHF and HS, consistent with healthy biomechanics \cite{Reznick2021}. For example, W-SD transitions showed reduced variance and better representation of human movement as presented in Fig.~\ref{fig:thighAngleTrajectories}, which is critical for FSM-based transition detection. The mapping also improved median W-SD accuracy by 20\% and maintained SA-W accuracy at 100\%, but slightly decreasing W-S and S-W accuracies by 10\%  as shown in Fig.~\ref{fig:RecognitionAccuracyTransitions2Exo_wandwoMapping}. Despite some declines, overall accuracy improved for many subjects with reduced variability across locomotion modes, showing the mapping function mitigates joint misalignment, enhancing locomotion detection and control. While challenges remain due to user- and system-specific variability, integrating this method into exoskeletons could provide more precise assistance.

\subsection{The Importance of Personalization}

Individual physiological characteristics and personalized interaction patterns significantly affect human-machine interaction and exoskeleton performance \cite{Kirkwood2021, Massardi2022}. As users exhibit distinct gait patterns, a strong model capacity to tackle specific cases is necessary. However, limited data availability and computational constraints often hinder the development of generalizable, efficient deep-learning models. In this context, we showed that lightweight ML-trained TH-methods combined with subject-specific optimization, enhance generalization and improve transition detection accuracy across locomotion modes. Results in Fig.~\ref{fig:MeanAccuracyChanges_ewalk_autonomyo_initialvsSBAvsBO} show SBA as an effective initial tuning method for simpler systems like eWalk but less effective for complex systems like autonomyo due to its one-step optimization approach. In contrast, BO achieves superior performance by iteratively refining parameters, and adapting to diverse devices and user variability. For example, BO improved eWalk transitions for S5 (W-SD and SD-W, from 20\% and 40\% to 100\%) and S2 (W-S and S-W, from 80\% and 60\% to 100\%). Similarly, BO significantly improved autonomyo transitions for S2 and 6 (e.g., W-S from 40\% and 60\% to 100\%). In addition, Fig.~\ref{fig:ThresholdChanges_Baseline_BO} highlights post-optimization thresholds diverging from baseline values, indicating that often optimal thresholds are unique to individuals with distinct gait patterns. These findings emphasize the necessity of personalized strategies and support a two-step optimization approach: SBA for initial simpler tuning, followed by BO for robust, adaptive performance if needed in the case of more complex systems such as autonomyo.

\begin{figure*}[t]
    \centering
    \includegraphics[width=1.0\linewidth]{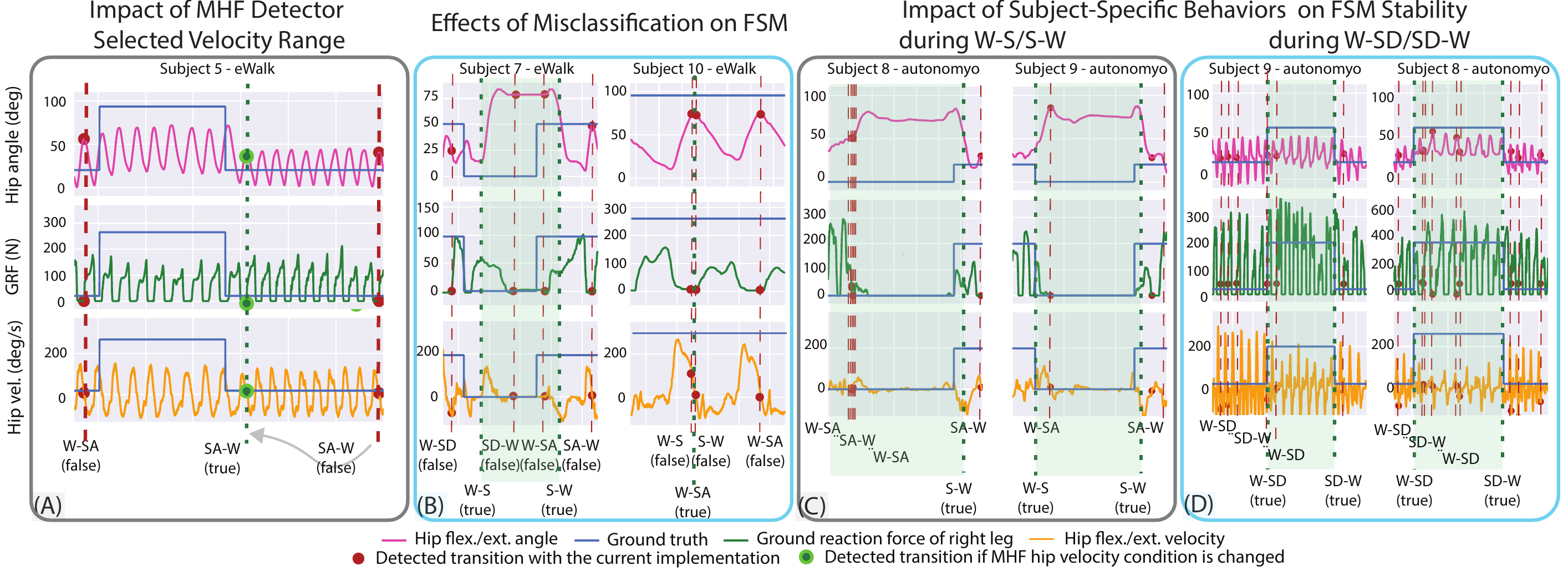}
    \caption{Investigating the false-detection of the transitions and the underlying reasons with main groups of MHF detector design, previous misclassification, and example of subject-specific patterns. Step changes in the GT signal indicate where the transition is happening.}
    \label{fig:allShowCases}
\end{figure*}

While BO optimization improved transition detection for most subjects as presented in Fig.~\ref{fig:SubjectSpecificChanges_ewalkandAutonomyo_v2}, some challenges remained due to non-optimized FSM components, subject-specific behaviors, and detector design issues.Concerning eWalk, S5's SA-W misclassifications were linked to the MHF detector design, which used a narrow velocity range near zero to detect the maximum in the hip signal. Expanding this range improved accuracy but revealed the detector’s limitations, as shown in Fig.\ref{fig:allShowCases}-A. S7 experienced FSM instability during W-S and S-W transitions due to FSM implementation rather than threshold issues, while S10 faced noise-induced misclassifications tied to the Loaded/Unloaded threshold, as outlined in Fig.\ref{fig:allShowCases}-B. In the case of autonomyo, S5, 8, and 9 encountered FSM instability during W-SD and SD-W transitions, often due to cautious, small steps or unique non-standard behaviors, such as high leg lifts before descending stairs in S8, as shown in Fig.~\ref{fig:allShowCases}-C/D, where multiple unstable detections are recognizable. During S9's W-S and S-W transitions, the system incorrectly detected SA, preventing the FSM from detecting the S-W transition, and highlighting issues beyond parameter optimization.

These challenges underscore that threshold optimization alone cannot address certain behavioral adaptations but also highlight the potential for further performance improvements if other parameters are considered while fine-tuning. Overall, controllers must balance responsiveness and stability, minimizing false positives and negatives to ensure safe and effective support. Ultimately, these findings indicate that a more flexible and adaptable exoskeleton design, combined with user training to interact effectively with the algorithm, is crucial for improving the control system's performance and enhancing both safety and effectiveness in human-machine interaction.

\subsection{The Importance of Real-Time Performance}

While personalization is key to improving exoskeleton performance it should also balance real-time computational demands. ML-trained TH-based method outperforms rule-based classifiers on embedded systems \cite{Orhan2023_2}, while deep learning, though viable on platforms like NVIDIA Jetson Xavier \cite{Vandersteegen2019}, is often constrained by specific hardware requirements (i.e. GPUs) and sampling rates \cite{Lu2020}.

In this study, we show that event-based state machines can be a computationally efficient alternative. SBA requires initial tuning but supports automated threshold adjustments, while BO eliminates expert input and shows to be four times faster than GSA as presented in Fig.~\ref{fig:CompTime_BO_GSA_ewalk}, achieving high performance at low computational cost, completing optimization in 2 minutes on a PC. Even with the limited computational power of an embedded system, similar outcomes could be achieved within a few minutes. 

\subsection{Limitations \& Future Work}

Despite the promising results, this study also reveals some limitations. A primary one is the absence of active assistance in the exoskeletons during testing, meaning user kinematics were unaffected by assistive torques. Expanding parameter optimization beyond thresholds could enhance performance further, and further research should be conducted in this sense. While BO showed strong results, it was not applied in real-time. The reliance on right-leg measurements risks misclassification for left-leg-dominant users; incorporating data from both legs could improve robustness. Future work should focus on large-scale, real-time parameter optimization and include participants with diverse characteristics and mobility impairments to evaluate performance across broader user profiles. 

\section{Conclusion}
\label{sec:Conclusion}

This study proposed two methods, the Statistics-Based Approach and Bayesian Optimization, to personalize ML-trained transition thresholds for lower-limb exoskeletons and addressed joint misalignment in human-exoskeleton interactions. Experiments with two exoskeletons, eWalk, and autonomyo, showed these methods effectively detected locomotion mode transitions by adapting to individual gait patterns and system-specific traits, ensuring reliable performance across users. An open-source dataset of lower-limb kinematics from 18 subjects with two exoskeletons is provided to support further research, and the results highlight the potential of proposed methods for assistive devices, where subject- and system-specific adaptation enhances transition detection. 

\vspace{-0.15cm}
\section*{Acknowledgment}
We thank Salim Boussofara for contributing to the dataset preparation and scripts, FAULHABER SA and Sonceboz for supporting exoskeleton development, and the subjects who participated in the experiments. We thank the vector drawings designed by stories/Freepik that have been used in Fig.\ref{fig:OnlineTestsOutdoorEnvironment}.

\vspace{-0.2cm}
\bibliographystyle{IEEEtran}
\bibliography{bibliography.bib}

\begin{thebibliography}{10}
\providecommand{\url}[1]{#1}
\csname url@rmstyle\endcsname
\providecommand{\newblock}{\relax}
\providecommand{\bibinfo}[2]{#2}
\providecommand\BIBentrySTDinterwordspacing{\spaceskip=0pt\relax}
\providecommand\BIBentryALTinterwordstretchfactor{4}
\providecommand\BIBentryALTinterwordspacing{\spaceskip=\fontdimen2\font plus
\BIBentryALTinterwordstretchfactor\fontdimen3\font minus \fontdimen4\font\relax}
\providecommand\BIBforeignlanguage[2]{{%
\expandafter\ifx\csname l@#1\endcsname\relax
\typeout{** WARNING: IEEEtran.bst: No hyphenation pattern has been}%
\typeout{** loaded for the language `#1'. Using the pattern for}%
\typeout{** the default language instead.}%
\else
\language=\csname l@#1\endcsname
\fi
#2}}

\bibitem{Kolaghassi2021}
R.~Kolaghassi, M.~K. Al-Hares, and K.~Sirlantzis, ``Systematic review of intelligent algorithms in gait analysis and prediction for lower limb robotic systems,'' \emph{IEEE Access}, vol.~9, pp. 113\,788--113\,812, 2021.

\bibitem{Donahue2022}
S.~R. Donahue and M.~E. Hahn, ``{Feature Identification With a Heuristic Algorithm and an Unsupervised Machine Learning Algorithm for Prior Knowledge of Gait Events},'' \emph{IEEE TNSRE}, vol.~30, pp. 108--114, 2022.

\bibitem{Gehlhar2023}
R.~Gehlhar, M.~Tucker, A.~J. Young, and A.~D. Ames, ``{A review of current state-of-the-art control methods for lower-limb powered prostheses},'' \emph{Annual Reviews in Control}, vol.~55, pp. 142--164, 2023.

\bibitem{Cheng2021}
S.~Cheng, E.~Bolivar-Nieto, and R.~D. Gregg, ``Real-time activity recognition with instantaneous characteristic features of thigh kinematics,'' \emph{IEEE TNSRE}, vol.~29, pp. 1827--1837, 2021.

\bibitem{Qian2022}
Y.~Qian, Y.~Wang, C.~Chen, J.~Xiong, Y.~Leng, H.~Yu, and C.~Fu, ``Predictive locomotion mode recognition and accurate gait phase estimation for hip exoskeleton on various terrains,'' \emph{IEEE RA-L}, vol.~7, no.~3, pp. 6439--6446, 2022.

\bibitem{Zhang2024}
X.~Zhang, E.~Tricomi, F.~Missiroli, N.~Lotti, and L.~Masia, ``Real-time assistive control via imu locomotion mode detection in a soft exosuit: An effective approach to enhance walking metabolic efficiency,'' \emph{IEEE/ASME Transactions on Mechatronics}, vol.~29, no.~3, pp. 1797--1808, 2024.

\bibitem{Haque2024}
M.~R. Haque, M.~R. Islam, E.~Sazonov, and X.~Shen, ``Swing-phase detection of locomotive mode transitions for smooth multi-functional robotic lower-limb prosthesis control,'' \emph{Frontiers in Robotics and AI}, vol.~11, 2024.

\bibitem{Wang2024}
Z.~Wang, J.~Pang, P.~Tao, Z.~Ji, J.~Chen, L.~Meng, R.~Xu, and D.~Ming, ``Locomotion transition prediction at anticipatory locomotor adjustment phase with shap feature selection,'' \emph{Biomedical Signal Processing and Control}, vol.~92, p. 106105, 2024.

\bibitem{Cheng_2024}
\BIBentryALTinterwordspacing
S.~Cheng, C.~A. Laubscher, and R.~D. Gregg, ``Controlling powered prosthesis kinematics over continuous inter-leg transitions between walking and stair ascent/descent,'' June 2024. [Online]. Available: \url{http://dx.doi.org/10.36227/techrxiv.171778818.88300126/v1}
\BIBentrySTDinterwordspacing

\bibitem{Zhou2019}
Z.~Zhou, X.~Liu, Y.~Jiang, J.~Mai, and Q.~Wang, ``{Real-time onboard SVM-based human locomotion recognition for a bionic knee exoskeleton on different terrains},'' in \emph{Wearable Robotics Association Conference (WearRAcon)}.\hskip 1em plus 0.5em minus 0.4em\relax IEEE, 2019, pp. 34--39.

\bibitem{Lu2020_2}
Z.~Lu, A.~Narayan, and H.~Yu, ``A deep learning based end-to-end locomotion mode detection method for lower limb wearable robot control,'' in \emph{IEEE IROS}, 2020, pp. 4091--4097.

\bibitem{Narayan2022}
A.~Narayan, F.~A. Reyes, M.~Ren, and Y.~Haoyong, ``{Real-Time Hierarchical Classification of Time Series Data for Locomotion Mode Detection},'' \emph{IEEE Journal of Biomedical and Health Informatics}, vol.~26, no.~4, pp. 1749--1760, 2022.

\bibitem{Vu2022}
H.~T.~T. Vu, H.-L. Cao, D.~Dong, T.~Verstraten, J.~Geeroms, and B.~Vanderborght, ``{Comparison of machine learning and deep learning-based methods for locomotion mode recognition using a single inertial measurement unit},'' \emph{Frontiers in Neurorobotics}, vol.~16, 2022.

\bibitem{Eken2023}
H.~Eken, F.~Lanotte, V.~Papapicco, M.~F. Penna, E.~Gruppioni, E.~Trigili, S.~Crea, and N.~Vitiello, ``A locomotion mode recognition algorithm using adaptive dynamic movement primitives,'' \emph{IEEE TNSRE}, vol.~31, pp. 4318--4328, 2023.

\bibitem{Young2014b}
A.~Young, A.~Simon, N.~Fey, and L.~Hargrove, ``Intent recognition in a powered lower limb prosthesis using time history information,'' \emph{Annals of Biomedical Engineering}, vol.~42, no.~3, pp. 631--641, 2014.

\bibitem{Su2019}
B.~Su, J.~Wang, S.~Liu, M.~Sheng, J.~Jiang, and K.~Xiang, ``A cnn-based method for intent recognition using inertial measurement units and intelligent lower limb prosthesis,'' \emph{IEEE TNSRE}, vol.~27, pp. 1032–--1042, 2019.

\bibitem{Moreira2022}
L.~Moreira, J.~Figueiredo, J.~Cerqueira, and C.~P. Santos, ``{A Review on Locomotion Mode Recognition and Prediction When Using Active Orthoses and Exoskeletons},'' \emph{Sensors}, vol.~22, no.~19, p. 7109, 2022.

\bibitem{Long2016}
Y.~Long, Z.-J. Du, W.-D. Wang, G.-Y. Zhao, G.-Q. Xu, L.~He, X.-W. Mao, and W.~Dong, ``Pso-svm-based online locomotion mode identification for rehabilitation robotic exoskeletons,'' \emph{Sensors}, vol.~16, no.~9, 2016.

\bibitem{Wang2018}
C.~Wang, X.~Wu, Y.~Ma, G.~Wu, and Y.~Luo, ``{A Flexible Lower Extremity Exoskeleton Robot with Deep Locomotion Mode Identification},'' \emph{Complexity}, vol. 2018, pp. 1--9, 2018.

\bibitem{Liu2020}
X.~Liu and Q.~Wang, ``{Real-Time Locomotion Mode Recognition and Assistive Torque Control for Unilateral Knee Exoskeleton on Different Terrains},'' \emph{IEEE/ASME Transactions on Mechatronics}, vol.~25, no.~6, pp. 2722--2732, 2020.

\bibitem{Du2021}
G.~Du, J.~Zeng, C.~Gong, and E.~Zheng, ``{Locomotion Mode Recognition with Inertial Signals for Hip Joint Exoskeleton},'' \emph{Applied Bionics and Biomechanics}, vol. 2021, pp. 1--11, 2021.

\bibitem{Kang2022}
I.~Kang, D.~D. Molinaro, G.~Choi, J.~Camargo, and A.~J. Young, ``{Subject-Independent Continuous Locomotion Mode Classification for Robotic Hip Exoskeleton Applications},'' \emph{IEEE Transactions on Biomedical Engineering}, vol.~69, no.~10, pp. 3234--3242, 2022.

\bibitem{Alzubaidi2023}
L.~Alzubaidi, J.~Bai, A.~Al-Sabaawi, J.~Santamaría, A.~S. Albahri, B.~S.~N. Al-dabbagh, M.~A. Fadhel, M.~Manoufali, J.~Zhang, A.~H. Al-Timemy, Y.~Duan, A.~Abdullah, L.~Farhan, Y.~Lu, A.~Gupta, F.~Albu, A.~Abbosh, and Y.~Gu, ``A survey on deep learning tools dealing with data scarcity: definitions, challenges, solutions, tips, and applications,'' \emph{Journal of Big Data}, vol.~10, no.~1, p.~46, 2023.

\bibitem{Orhan2023_2}
Z.~Ã. Orhan, A.~Dal~Prete, A.~Bolotnikova, M.~Gandolla, A.~Ijspeert, and M.~Bouri, ``Real-time locomotion transitions detection: Maximizing performances with minimal resources,'' in \emph{IEEE ICRA}, 2024, pp. 3241--3247.

\bibitem{Quesada2016}
R.~E. Quesada, J.~M. Caputo, and S.~H. Collins, ``Increasing ankle push-off work with a powered prosthesis does not necessarily reduce metabolic rate for transtibial amputees,'' \emph{Journal of Biomechanics}, vol.~49, no.~14, pp. 3452--3459, 2016.

\bibitem{Kim2017_2}
M.~Kim, Y.~Ding, P.~Malcolm, J.~Speeckaert, C.~J. Siviy, C.~J. Walsh, and S.~Kuindersma, ``Human-in-the-loop bayesian optimization of wearable device parameters,'' \emph{PLOS ONE}, vol.~12, no.~9, pp. 1--15, 09 2017.

\bibitem{Slade2024}
P.~Slade, C.~Atkeson, J.~M. Donelan, H.~Houdijk, K.~A. Ingraham, M.~Kim, K.~Kong, K.~L. Poggensee, R.~Riener, M.~Steinert, J.~Zhang, and S.~H. Collins, ``{On human-in-the-loop optimization of human–robot interaction},'' \emph{Nature}, vol. 633, no. 8031, pp. 779--788, sep 2024.

\bibitem{Zhang2017}
J.~Zhang, P.~Fiers, K.~A. Witte, R.~W. Jackson, K.~L. Poggensee, C.~G. Atkeson, and S.~H. Collins, ``Human-in-the-loop optimization of exoskeleton assistance during walking,'' \emph{Science}, vol. 356, no. 6344, pp. 1280--1284, 2017.

\bibitem{Ding2018}
Y.~Ding, M.~Kim, S.~Kuindersma, and C.~J. Walsh, ``Human-in-the-loop optimization of hip assistance with a soft exosuit during walking,'' \emph{Science Robotics}, vol.~3, no.~15, p. eaar5438, 2018.

\bibitem{Xu2023}
L.~Xu, X.~Liu, Y.~Chen, L.~Yu, Z.~Yan, C.~Yang, C.~Zhou, and W.~Yang, ``Reducing the muscle activity of walking using a portable hip exoskeleton based on human-in-the-loop optimization,'' \emph{Frontiers in Bioengineering and Biotechnology}, vol.~11, 2023.

\bibitem{Brochu2010}
\BIBentryALTinterwordspacing
E.~Brochu, V.~M. Cora, and N.~de~Freitas, ``A tutorial on bayesian optimization of expensive cost functions, with application to active user modeling and hierarchical reinforcement learning,'' \emph{CoRR}, vol. abs/1012.2599, 2010. [Online]. Available: \url{http://arxiv.org/abs/1012.2599}
\BIBentrySTDinterwordspacing

\bibitem{Wen2020}
T.-C. Wen, M.~Jacobson, X.~Zhou, H.-J. Chung, and M.~Kim, ``The personalization of stiffness for an ankle-foot prosthesis emulator using human-in-the-loop optimization,'' in \emph{IEEE/RSJ IROS}, 2020, pp. 3431--3436.

\bibitem{Reznick2021}
E.~Reznick, K.~R. Embry, R.~Neuman, E.~Bolivar-Nieto, and R.~D. Gregg, ``Lower-limb kinematics and kinetics during continuously varying human locomotion,'' \emph{Nature, Scientific Data}, vol.~8, 2021.

\bibitem{Scikit2011}
F.~Pedregosa~et al, ``Scikit-learn: Machine learning in python,'' \emph{JMLR 12, pp.}, 2011.

\bibitem{Messara2023}
S.~Messara, A.~R. Manzoori, A.~Di~Russo, A.~Ijspeert, and M.~Bouri, ``Novel design and implementation of a neuromuscular controller on a hip exoskeleton for partial gait assistance,'' in \emph{IEEE ICORR}, 2023, pp. 1--6.

\bibitem{Ortlieb2017}
A.~Ortlieb, M.~Bouri, R.~Baud, and H.~Bleuler, ``An assistive lower limb exoskeleton for people with neurological gait disorders,'' in \emph{IEEE ICORR}, 2017, pp. 441--446.

\bibitem{orhan2023}
Z.~Ã. Orhan, M.~Shafiee, V.~Juillard, J.~C. Oliveira, A.~Ijspeert, and M.~Bouri, ``Exorecovery: Push recovery with a lower-limb exoskeleton based on stepping strategy,'' in \emph{IEEE ICRA}, 2024, pp. 3248--3255.

\bibitem{Schiele2006}
A.~Schiele and F.~C. van~der Helm, ``Kinematic design to improve ergonomics in human machine interaction,'' \emph{IEEE TNSRE : a publication of the IEEE Engineering in Medicine and Biology Society}, vol. 14(4), pp. 456--469, 2006.

\bibitem{Stienen2009}
A.~Stienen, E.~Hekman, F.~van~der Helm, and H.~Kooij, ``Self-aligning exoskeleton axes through decoupling of joint rotations and translations,'' \emph{Robotics, IEEE Transactions on}, vol.~25, pp. 628 -- 633, 07 2009.

\bibitem{Celebi2013}
B.~Celebi, M.~Yalcin, and V.~Patoglu, ``Assiston-knee: A self-aligning knee exoskeleton,'' in \emph{IEEE IROS}, 2013, pp. 996--1002.

\bibitem{Gálvez2016}
M.~Gálvez and A.~Aceves-Lopez, ``A review on compliant joint mechanisms for lower limb exoskeletons,'' \emph{Journal of Robotics}, 2016.

\bibitem{Garnett_2023}
R.~Garnett, \emph{Bayesian Optimization}.\hskip 1em plus 0.5em minus 0.4em\relax Cambridge University Press, 2023.

\bibitem{GPy}
{GPy}, ``{GPy}: A gaussian process framework in python,'' \url{http://github.com/SheffieldML/GPy}, since 2012.

\bibitem{Rasmussen2006Gaussian}
C.~E. Rasmussen and C.~K.~I. Williams, \emph{Gaussian Processes for Machine Learning}.\hskip 1em plus 0.5em minus 0.4em\relax The MIT Press, 2006.

\bibitem{Chollet2021}
F.~Chollet, \emph{Deep Learning with Python, Second Edition}.\hskip 1em plus 0.5em minus 0.4em\relax Manning, 2021.

\bibitem{dataset2024}
``Exomove - biomechanical dataset of daily activities with lower-limb exoskeletons,'' \url{https://zenodo.org/records/14076104}, accessed: 2024-11-14.

\bibitem{Moore2015}
J.~K. Moore, S.~K. Hnat, and A.~J. van~den Bogert, ``An elaborate data set on human gait and the effect of mechanical perturbations,'' \emph{PeerJ}, vol.~3, p. e918, 2015.

\bibitem{Fukuchi2018}
C.~A. Fukuchi, R.~Fukuchi, and M.~Duarte, ``A public dataset of overground and treadmill walking kinematics and kinetics in healthy individuals,'' \emph{PeerJ}, vol.~6, p. e4640, 2018.

\bibitem{Hu2018}
B.~Hu, E.~Rouse, and L.~Hargrove, ``Benchmark datasets for bilateral lower-limb neuromechanical signals from wearable sensors during unassisted locomotion in able-bodied individuals,'' \emph{Frontiers in Robotics and AI}, vol.~5, p.~14, 2018.

\bibitem{Camargo2021}
J.~Camargo, A.~Ramanathan, W.~Flanagan, and A.~J. Young, ``A comprehensive open-source dataset of lower limb biomechanics in multiple conditions of stairs, ramps, and level-ground ambulation and transitions,'' \emph{Elsevier, Journal of Biomechanics}, vol. 119, 2021.

\bibitem{Naf2019}
M.~B. Näf, K.~Junius, M.~Rossini, C.~Rodriguez-Guerrero, B.~Vanderborght, and D.~Lefeber, ``{Misalignment Compensation for Full Human-Exoskeleton Kinematic Compatibility: State of the Art and Evaluation},'' \emph{Applied Mechanics Reviews}, vol.~70, no.~5, p. 050802, 02 2019.

\bibitem{Siviy2023}
C.~Siviy, L.~M. Baker, B.~T. Quinlivan, F.~Porciuncula, K.~Swaminathan, L.~N. Awad, and C.~J. Walsh, ``Opportunities and challenges in the development of exoskeletons for locomotor assistance,'' \emph{Nature Biomedical Engineering}, vol.~7, no.~4, pp. 456--472, 2023.

\bibitem{BesslerEtten2022}
J.~Bessler-Etten, L.~Schaake, G.~B. Prange-Lasonder, and J.~H. Buurke, ``Assessing effects of exoskeleton misalignment on knee joint load during swing using an instrumented leg simulator,'' \emph{Journal of NeuroEngineering and Rehabilitation}, vol.~19, no.~1, p.~13, 2022.

\bibitem{Kirkwood2021}
G.~L. Kirkwood, C.~D. Otmar, and M.~Hansia, ``Who's leading this dance?: Theorizing automatic and strategic synchrony in human-exoskeleton interactions,'' \emph{Frontiers in Psychology}, vol.~12, 2021.

\bibitem{Massardi2022}
S.~Massardi, D.~Rodriguez-Cianca, D.~Pinto-Fernandez, J.~C. Moreno, M.~Lancini, and D.~Torricelli, ``Characterization and evaluation of human–exoskeleton interaction dynamics: A review,'' \emph{Sensors}, vol.~22, no.~11, 2022.

\bibitem{Vandersteegen2019}
M.~Vandersteegen, K.~V. Beeck, and T.~Goedem{\'e}, ``Super accurate low latency object detection on a surveillance uav,'' \emph{International Conference on Machine Vision Applications}, pp. 1--6, 2019.

\bibitem{Lu2020}
H.~Lu, L.~R. Schomaker, and R.~Carloni, ``Imu-based deep neural networks for locomotor intention prediction,'' in \emph{IEEE IROS}, 2020, pp. 4134--4139.

\end{thebibliography}

\vspace{-1.5cm}
\begin{IEEEbiography}[{\includegraphics[width=76.5in,height=0.9in,clip,keepaspectratio]{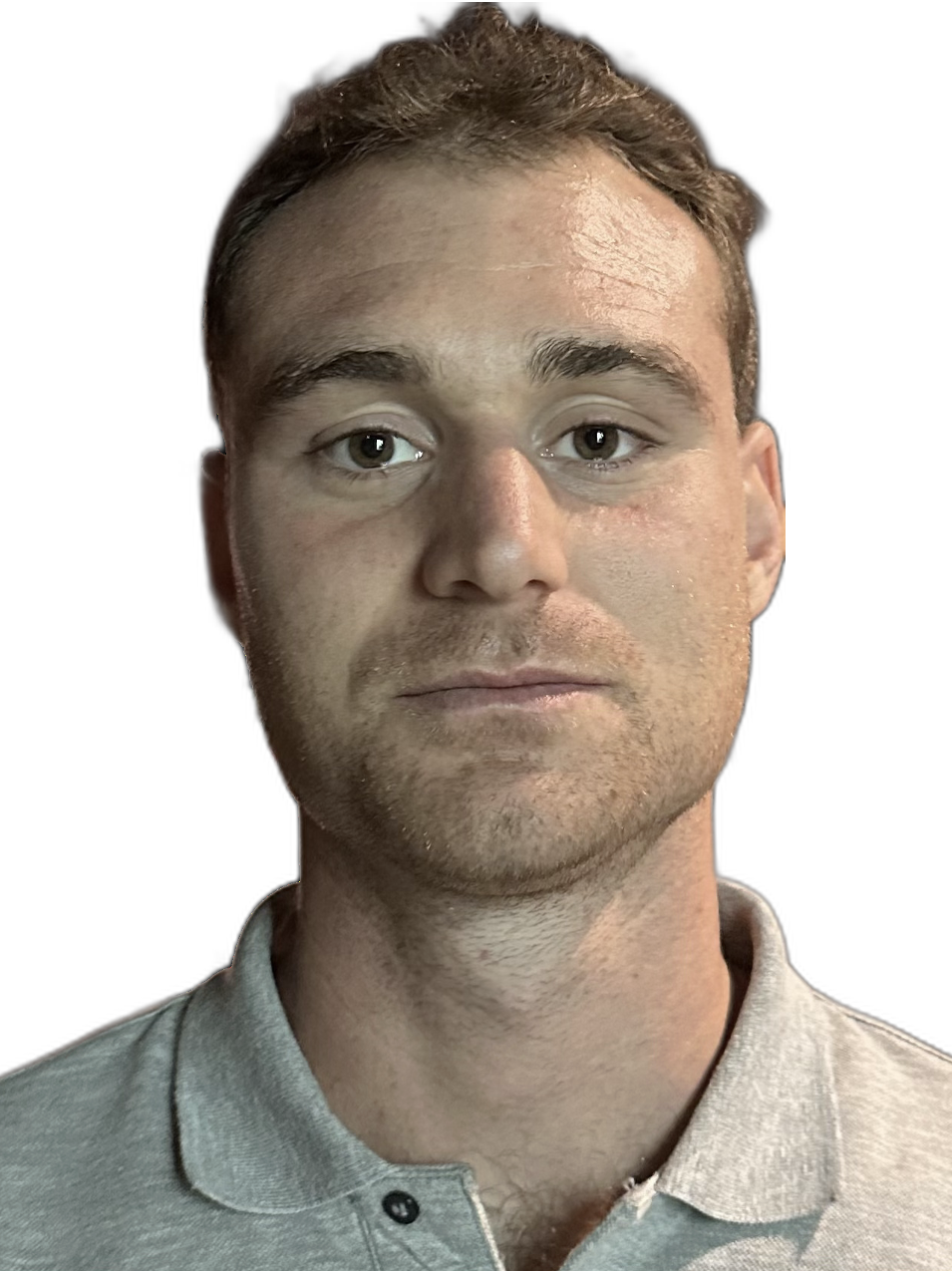}}]
{Andrea Dal Prete} received his B.S. degree in Mechanical Engineering in 2021 and his M.S. degree in Mechatronics Engineering in 2023, from Politecnico di Milano, Italy. Currently, he is a Ph.D. candidate in Mechanical Engineering at Politecnico di Milano. His research focuses on bioinspired and intelligent robotics and motor assistance for industrial activities, particularly for workers performing exhausting tasks.
\end{IEEEbiography}

\vspace{-1.5cm}
\begin{IEEEbiography}[{\includegraphics[width=76.5in,height=0.9in,clip,keepaspectratio]{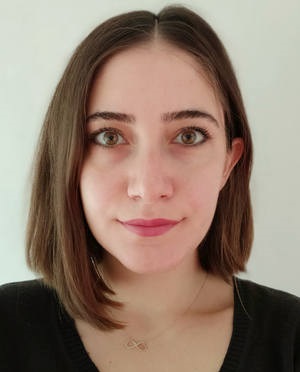}}]
{Zeynep {\"O}zge Orhan} received her B.S. degree in Mechanical Engineering from METU, Turkey, in 2017, and her M.S. degree in Mechatronics Engineering from Sabanci University, Turkey, in 2020. She is currently a Ph.D. candidate at EPFL. Her research focuses on developing assistive control strategies for lower-limb exoskeletons to assist daily living activities.
\end{IEEEbiography}

\vspace{-1.5cm}
\begin{IEEEbiography}[{\includegraphics[width=76.5in,height=0.9in,clip,keepaspectratio]{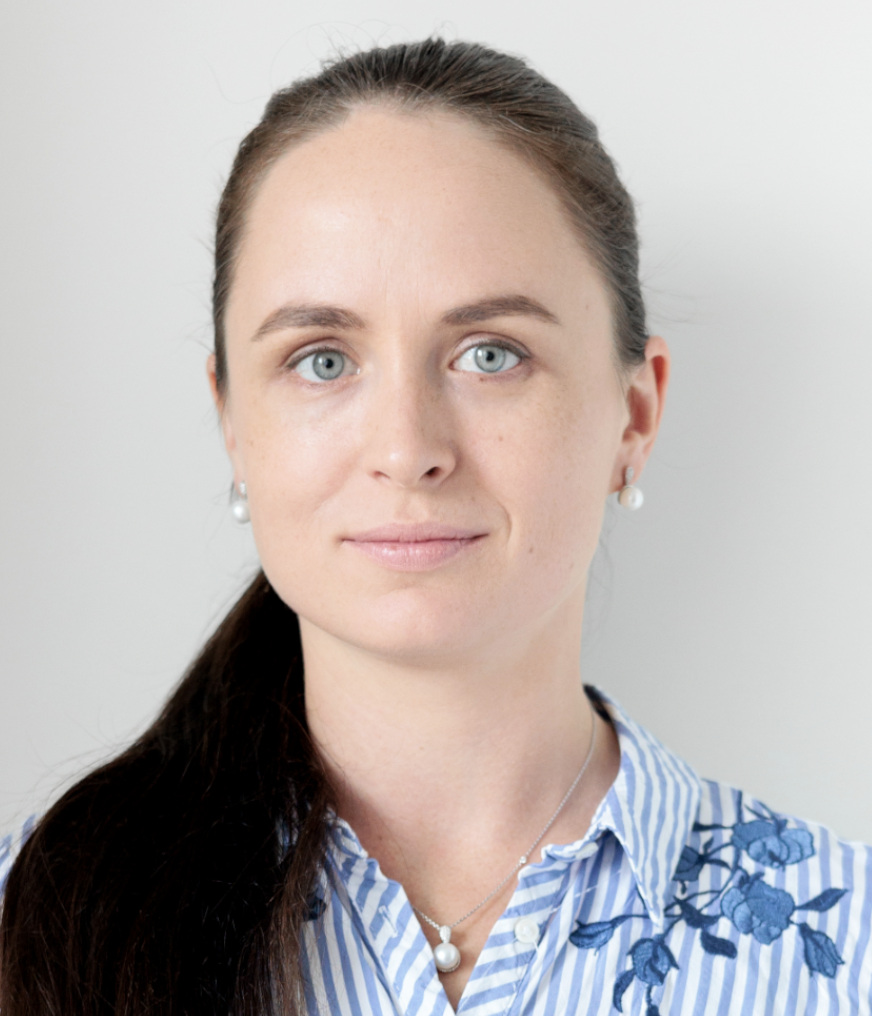}}]
{Anastasia Bolotnikova} received her M.Sc. degree in Computer Science from the University of Tartu, Estonia, in 2017 and her Ph.D. in Robotics at the University of Montpellier, France, in 2021, then she was a postdoctoral researcher at EPFL. Since 2024, she is a research fellow at the LAAS-CNRS. Her research interests include robotic physical assistance and tactile communication.
\end{IEEEbiography}

\vspace{-1.5cm}
\begin{IEEEbiography}[{\includegraphics[width=76.5in,height=0.9in,clip,keepaspectratio]{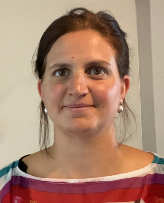}}]
{Marta Gandolla} is an Assistant Professor in the Department of Mechanical Engineering at Politecnico di Milano. She received her M.Sc. in Biomedical Engineering in 2009 and her Ph.D. in Bioengineering in 2013 from Politecnico di Milano. Her research explores neurological motor rehabilitation and motor assistance, with a focus on daily life activities for fragile individuals and workers in physically demanding tasks. 
\end{IEEEbiography}

\vspace{-1.5cm}
\begin{IEEEbiography}[{\includegraphics[width=76.5in,height=0.9in,clip,keepaspectratio]{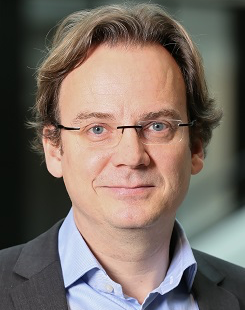}}]
{Auke Ijspeert} is a Full Professor at EPFL and head of the Biorobotics Laboratory. He holds a B.Sc. and M.Sc. in Physics from EPFL (1995) and a Ph.D. in Artificial Intelligence from the University of Edinburgh (1999). His research lies at the intersection of robotics, computational neuroscience, nonlinear dynamical systems, and applied machine learning. He also works on assisting people with limited mobility using exoskeletons and assistive furniture.
\end{IEEEbiography}

\vspace{-1.5cm}
\begin{IEEEbiography}[{\includegraphics[width=76.5in,height=0.9in,clip,keepaspectratio]{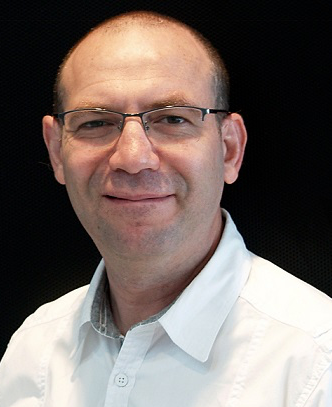}}]
{Mohamed Bouri} is the group leader of the REHAssist group at EPFL. He received his degree in Electrical Engineering in 1992 and his Ph.D. in Industrial Automation in 1997 from INSA Lyon, France. Since joining EPFL in 1997, he has been active in robot control, automation, and design for medical and industrial applications. His current research focuses on exoskeleton development and associated control strategies. 
\end{IEEEbiography}

\end{document}